\documentclass{article}

\PassOptionsToPackage{numbers,compress}{natbib}
\usepackage[final,main]{neurips_2026}

\usepackage[utf8]{inputenc}
\usepackage[T1]{fontenc}
\usepackage{hyperref}
\usepackage{url}
\usepackage{booktabs}
\usepackage{amsfonts}
\usepackage{amsmath}
\usepackage{nicefrac}
\usepackage{microtype}
\usepackage{bbm}
\usepackage{xcolor}
\usepackage{graphicx}
\usepackage{cleveref}
\usepackage{enumitem}
\usepackage{multirow}
\usepackage{subcaption}
\usepackage{algorithm}
\usepackage{algorithmic}
\usepackage{array}       
\usepackage{makecell}    
\usepackage{bm}        
\usepackage{mathtools}
\usepackage{adjustbox}
\usepackage{amssymb}
\usepackage{colortbl} 
\definecolor{lightblue}{RGB}{55, 140, 255}

\newcommand{\tworow}[2]{\begin{tabular}[c]{@{}c@{}}#1\vspace{-2pt}\\#2\end{tabular}}

\title{Rebalancing Reference Frame Dominance \\ to Improve Motion in Image-to-Video Models}
\author{%
  \textbf{Wooseok Jeon}\textsuperscript{1}\thanks{Equal contribution.}
  \quad \textbf{Seungho Park}\textsuperscript{1}\footnotemark[1]
  \quad \textbf{Seunghyun Shin}\textsuperscript{2}
  \quad \textbf{Sangeyl Lee}\textsuperscript{1}
  \\[2pt]
  \textbf{Hyeonho Jeong}\textsuperscript{\textbf{3}}
  \quad \textbf{Hae-Gon Jeon}\textsuperscript{\textbf{1}}\thanks{Corresponding author.}
  \\[2pt]
  \textsuperscript{1}Yonsei University
  \quad \textsuperscript{2}GIST
  \quad \textsuperscript{3}Adobe Research
}

\begin{document}

\maketitle
\setcounter{footnote}{0}

\begin{figure*}[h]
\vspace{-7mm}
\centering
\includegraphics[width=\linewidth]{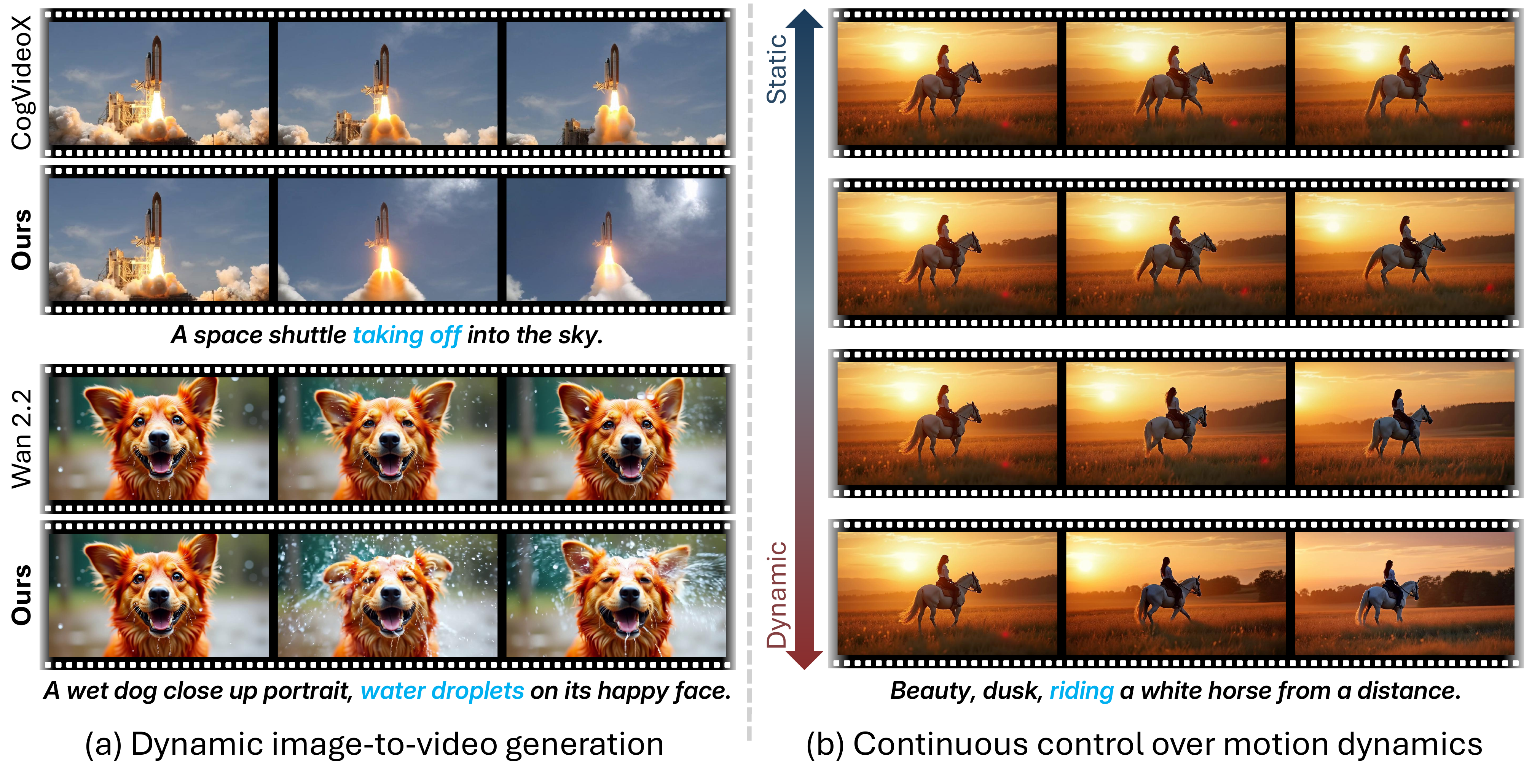}
\caption{\textbf{Example videos from our method.} We present DyMoS, a training-free and model-agnostic method for improving motion dynamics in image-to-video generation. (a)~Comparison of generated videos from the same input image. DyMoS produces dynamic motion while preserving video quality. (b)~Furthermore, DyMoS provides continuous control over motion dynamics.}
\label{fig1}
\vspace{-1mm}
\end{figure*}

\begin{abstract}
Image-to-video models often generate videos that remain overly static compared to text-to-video models. 
While prior approaches mitigate this issue by weakening or modifying the image-conditioning signal, they often require additional training or sacrifice fidelity to the reference image. 
In this work, we identify \emph{reference-frame dominance} as a key mechanism behind motion suppression.
We observe that non-reference frames in I2V models allocate excessive self-attention to reference-frame key tokens, causing reference information to be over-propagated across time and suppressing inter-frame dynamics. 
Based on this finding, we propose DyMoS~(Dynamic Motion Slider), a training-free and model-agnostic method that rebalances the attention pathway from generated frames to the reference frame during initial denoising steps. 
DyMoS leaves both the input image and model weights unchanged and introduces a single scalar parameter for continuous control over motion strength. 
Experiments across multiple state-of-the-art I2V backbones demonstrate that DyMoS consistently improves motion dynamics while maintaining visual quality and fidelity to the reference image. Project page: {\fontfamily{zi4}\selectfont \textcolor{lightblue}{\url{https://i2v-dymos.github.io/}}}
\end{abstract}
\section{Introduction}
\label{sec:intro}
Recent advances in generative models have been driven by the remarkable success of diffusion~\citep{ho2020denoising, song2021scorebased, sohl2015deep, dhariwal2021diffusion} 
and flow-matching~\citep{liu2022flow, lipman2023flow, albergo2023building, esser2024scaling} frameworks. Building on these paradigms, text-to-video~(T2V) models~\citep{hunyuanvideo2025, wan2025wan, yang2025cogvideox, blattmann2023stable, ho2022video, lin2024open, ho2022imagen} 
have made it possible to generate high-quality videos from text prompts.
Image-to-video~(I2V) models~\citep{hunyuanvideo2025, wan2025wan, yang2025cogvideox, blattmann2023stable, guo2024animatediff,hacohen2024ltx, ni2023conditional, shi2024motion, teng2025magi, xing2024dynamicrafter, zhang2023i2vgen, jeong2025track4gen} extend this capability by conditioning generation on reference images\footnote{In this paper, a reference image serves as the initial frame of the generated video, rather than functioning solely as an appearance or identity reference.}, enabling the creation of videos that follow text prompts while preserving the content and appearance of the input. 
These models are typically adapted from large-scale T2V backbones by introducing an additional image-conditioning pathway, which grounds the generation in specific visual input while retaining the motion priors learned from large-scale video data.

Despite this progress, I2V models often produce videos that remain overly static compared to their T2V counterparts, even when the prompt explicitly describes motion. 
Prior works address this static motion bias by either modifying the image-conditioning pathway or weakening the conditioning signal during generation~\citep{zhao2024identifying, ge2025flashi2v, choi2025enhancing}. 
While these approaches can encourage stronger motion, they often require additional training or involve direct modification of the input signal.
In the latter case, improved motion dynamics are achieved at the cost of fidelity to the reference image.

In this work, we revisit the static motion bias by analyzing how reference image information is utilized within I2V models.
Since I2V models process spatiotemporal latent tokens through self-attention, their attention maps allow us to analyze how information flows across frames during the iterative denoising process. 
By comparing paired T2V and I2V generations, we observe that query tokens in non-reference frames in I2V models allocate substantially more attention to the reference-frame key tokens than their paired T2V counterparts.
We refer to this asymmetric attention pattern as \emph{reference-frame dominance}.

This observation implies that the static nature of I2V models may stem from excessive propagation of reference-frame information through internal attention routing, rather than that from the input image itself.
To test this hypothesis, we directly modulate the attention logits associated with reference-frame key tokens.
Strengthening these logits makes generation more static, whereas weakening them restores motion and reduces the reference-frame dominance.
This provides intervention-based evidence that the reference-frame dominance is a primary driver of motion suppression in I2V models.

Based on this insight, we present DyMoS~(Dynamic Motion Slider), a simple, training-free method that mitigates the reference-frame dominance by rebalancing attention logits. 
DyMoS introduces a scalar bias shift to the attention logits corresponding to reference-frame key tokens during initial denoising steps, leaving the input image and model weights unchanged. 
This allows DyMoS to improve motion dynamics while preserving fidelity to the reference image. 
The magnitude of this bias serves as a single parameter for continuously controlling the generation, ranging from strict reference-image preservation to highly dynamic motion.

We validate the effectiveness of DyMoS in improving motion dynamics across state-of-the-art I2V models, including Wan 2.2~\citep{wan2025wan}, Wan 2.1~\citep{wan2025wan}, HunyuanVideo-1.5~\citep{hunyuanvideo2025} and CogVideoX-5B~\citep{yang2025cogvideox}. 
Across all backbones, DyMoS consistently improves motion dynamics while preserving reference fidelity, incurring negligible additional inference-time overhead (e.g., 1.2\% on Wan 2.2).
We further show that DyMoS provides continuous control over the static-to-dynamic spectrum, enabling users to adjust motion strength with a single scalar parameter. 
These results demonstrate that static motion bias in I2V generation can be effectively mitigated by intervening at the attention-logit level, without fine-tuning the model weights or degrading the input image.
\section{Background}
\label{sec:prelim}

\paragraph{Text-to-video models.} Modern flow-based text-to-video (T2V) models~\citep{wan2025wan, hunyuanvideo2025} employ diffusion transformers (DiTs)~\citep{peebles2023scalable, ma2024sit}, operating in the latent space of a 3D variational autoencoder (VAE)~\citep{kingma2013auto}, which consists of an encoder $\mathcal{E}(\cdot)$ and a decoder $\mathcal{D}(\cdot)$.
An RGB video $\bm{x}$ is first compressed into a low-dimensional latent representation via $\bm{z}_0=\mathcal{E}(\bm{x})$.
The latent $\bm{z}_0$ is then perturbed along a flow-matching interpolation path $\bm{z}_t=(1-t)\bm{z}_0+t\bm{\epsilon}$, where $t\in[0,1]$ and $\bm{\epsilon}\sim\mathcal{N}(\bm{0},\bm{I})$.
A neural network $\bm{v}_\theta(\bm{z}_t,t,\bm{c})$ is trained to approximate a target velocity $\bm{v}=\bm{\epsilon}-\bm{z}_0$, where $\bm{c}$ denotes the text condition embedded by a pretrained text encoder such as T5~\citep{raffel2020exploring}. Generation starts from a Gaussian latent $\bm{z}_T \sim \mathcal{N}(\bm{0}, \bm{I})$ and progressively denoises it into a clean latent $\bm{z}_0$ using the learned velocity field. The final latent is then decoded into pixel space as $\hat{\bm{x}}=\mathcal{D}(\bm{z}_0)$.

During the denoising process, classifier-free guidance (CFG)~\citep{ho2022classifier} is commonly applied in order to improve text-conditional generation by extrapolating between the unconditional and conditional predictions:
\begin{equation}
\label{eq:1}
\tilde{\bm{v}}_\theta(\bm{z}_t,t,\bm{c})
=
\bm{v}_\theta(\bm{z}_t,t,\varnothing)
+
\omega\left(
\bm{v}_\theta(\bm{z}_t,t,\bm{c})
-
\bm{v}_\theta(\bm{z}_t,t,\varnothing)
\right),
\end{equation}
where $\omega\geq1$ is the guidance scale, and $\varnothing$ denotes the null text condition. 

\paragraph{Self-attention mechanism.} Within each DiT block, spatiotemporal latent tokens are processed by self-attention. Let $S=f\cdot h\cdot w$ be the number of latent tokens. 
For each attention head, the attention weights are computed as
\begin{equation}
    \mathcal{A} = \operatorname{softmax}(\mathcal{L}), 
    \qquad
    \mathcal{L} = \frac{QK^\top}{\sqrt{d_h}}\in\mathbb{R}^{S\times S},
\end{equation}
where $Q,K\in\mathbb{R}^{S\times d_h}$  are query and key projections, $d_h$ is the per-head dimension, and $\mathcal{L}$ denotes the attention logits.
Since the token sequence spans both spatial and temporal dimensions, this operation aggregates information within and across frames.

\paragraph{Image-to-video models.} Image-to-video (I2V) models are commonly obtained by fine-tuning T2V models to additionally condition on a reference image $\bm{x}_{\mathrm{ref}}$, which corresponds to the first frame of the generated video. 
The reference image is first encoded as $\mathcal{E}(\bm{x}_{\mathrm{ref}})$ and temporally aligned with the noisy video latent.
For simplicity, we denote the resulting reference-image latent by $\bm{z}_{\mathrm{ref}}$.
The aligned reference-image latent is then concatenated with the noisy video latent along the channel dimension.
This gives the denoising network direct access to reference-image features throughout sampling.
During inference with an I2V model, CFG in Eq.~\ref{eq:1} is extended by providing the reference-image latent to both predictions as
\begin{equation}
\label{eq:3}
\tilde{\bm{v}}_{\theta}
(\bm{z}_t,t,\bm{c},\bm{z}_{\mathrm{ref}})
=
\bm{v}_\theta(\bm{z}_t,t,\varnothing,\bm{z}_{\mathrm{ref}})
+
\omega\left(
\bm{v}_\theta(\bm{z}_t,t,\bm{c},\bm{z}_{\mathrm{ref}})
-
\bm{v}_\theta(\bm{z}_t,t,\varnothing,\bm{z}_{\mathrm{ref}})
\right).
\end{equation}

Some I2V models further preserve the reference frame by replacing the corresponding latent with the encoded reference latent after each sampling update. 

\paragraph{Static motion bias in I2V generation.}
Although I2V models can faithfully preserve the appearance of the reference image, they often suffer from a stronger static motion bias than their T2V counterparts, where excessive reliance on the input image leads to videos with weak or
limited motion. Several recent methods~\citep{zhao2024identifying, ge2025flashi2v, choi2025enhancing, song2025historyguided} mitigate this bias by weakening or redesigning the image-conditioning pathway.
\citet{zhao2024identifying} attribute this phenomenon to conditional image leakage and mitigate it by perturbing the conditioning image with time-dependent noise, thereby reducing the model's reliance on the reference signal. FlashI2V~\citep{ge2025flashi2v} also targets the conditional
image leakage by reformulating flow matching with Fourier-guided latent shifting, which injects the image condition implicitly rather than providing the reference latent through direct concatenation.
Most closely related to our work, ALG~\citep{choi2025enhancing} addresses over-conditioning at inference time by suppressing high-frequency components of the input image during early denoising steps.

However, these methods either require additional training or introduce a trade-off between motion enhancement and reference-image preservation. 
In contrast, our method leaves the image condition unchanged and directly intervenes in the self-attention mechanism during inference, enabling training-free and model-agnostic control over motion dynamics while preserving fidelity to the input image.

\section{DyMoS: Dynamic Motion Slider}
\label{sec:method} 
In this section, we first analyze frame-to-frame self-attention in I2V generation
(\S\ref{sec:sa_comparison}) and examine whether modulating this attention pattern controls motion dynamics
(\S\ref{sec:gamma_sweep}). Finally, we present DyMoS, a training-free
and model-agnostic self-attention modulation method
(\S\ref{sec:DyMoS}).

\begin{figure*}[t]
\centering
\includegraphics[width=\linewidth]{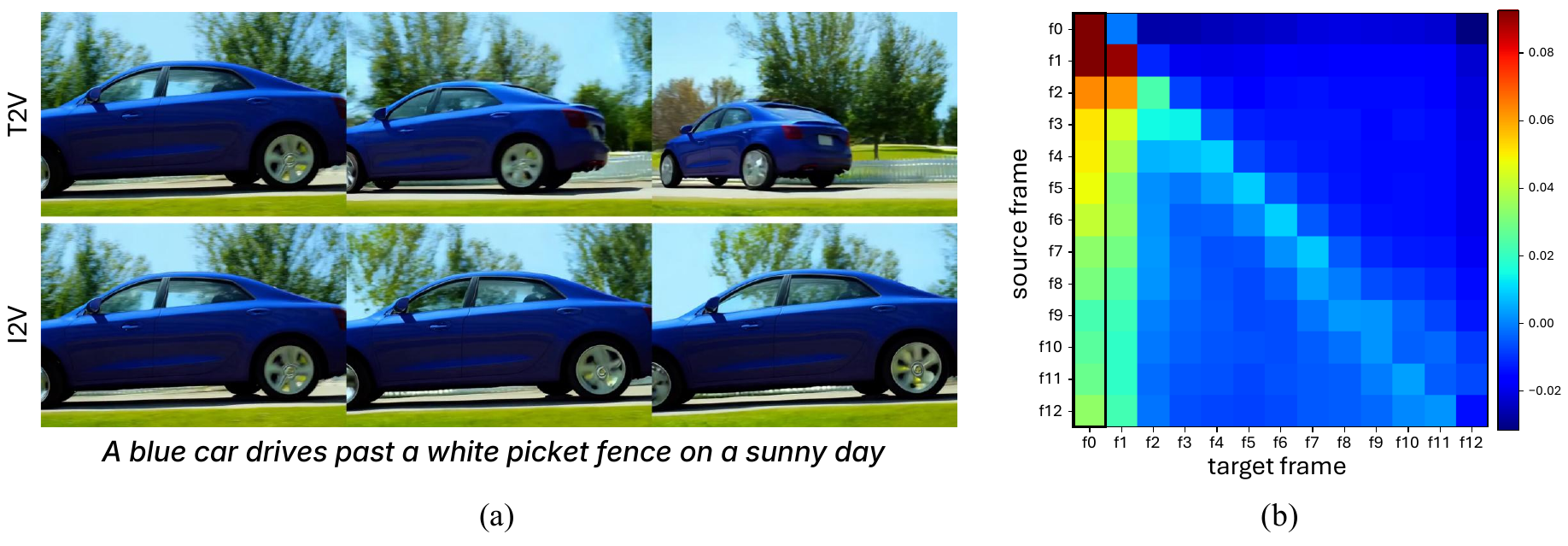}
\caption{
\textbf{Reference-frame dominance in I2V self-attention.}
(a) Qualitative comparison between paired T2V and I2V generations.
(b) Frame-to-frame self-attention difference map
$\mathcal{A}^{\mathrm{I2V}}-\mathcal{A}^{\mathrm{T2V}}$, averaged over the first 10\% of inference steps.}
\vspace{-3mm}
\label{fig2}
\end{figure*}

\subsection{Reference-frame dominance in I2V self-attention} 
\label{sec:sa_comparison} 

To understand how image conditioning changes information routing within the model, we compare frame-to-frame self-attention distributions between an I2V model and its corresponding T2V counterpart. 
We first generate 50 videos using a T2V model, and then use the first frames as reference images for an I2V model, using the same text prompts sourced from T2V-CompBench~\citep{sun2025t2v}. 
Since this setup ensures the same prompt and first frame for both models, we can analyze the differences in their respective attention distributions. 
The detailed procedure is provided in Appendix~\ref{app:sa_analysis_details}. 

In this paired setting, the two models generate videos with distinct motion dynamics and self-attention patterns.
As shown in Fig.~\ref{fig2}(a), the I2V model tends to preserve the reference image with limited inter-frame movement, whereas the T2V counterpart exhibits more dynamic motion given the same prompt. Fig.~\ref{fig2}(b) shows the frame-to-frame attention difference map $\mathcal{A}^{\mathrm{I2V}}-\mathcal{A}^{\mathrm{T2V}}$.
Positive values indicate that I2V allocates significantly more attention to the reference key tokens than its T2V counterpart as highlighted in black.
We refer to this structural asymmetry as \emph{reference-frame dominance}. 
Notably, this dominance is most pronounced during the initial denoising stage, which is consistent with the coarse-to-fine nature of the denoising process.
Since early steps largely determine global motion of the generated video~\citep{choi2025enhancing, kim2026motioncfg, jeon2026motion}, we target our subsequent intervention at this stage.

\subsection{Correlation between reference-frame dominance and motion dynamics}
\label{sec:gamma_sweep}
Based on this observation, we hypothesize that reference-frame dominance is a primary factor underlying static I2V generation, and we demonstrate that directly modulating this dominance impacts the motion dynamics of the generated videos.
Since the reference frame contains the input image information, excessive attention to reference-frame key tokens propagates the reference appearance along the temporal axis, thereby weakening frame-to-frame interactions necessary for motion dynamics. 
Accordingly, we argue that attenuating reference-frame dominance should affect both the generated motion and the frame-to-frame attention structure.

\begin{figure*}[t]
  \centering
  \includegraphics[width=\linewidth]{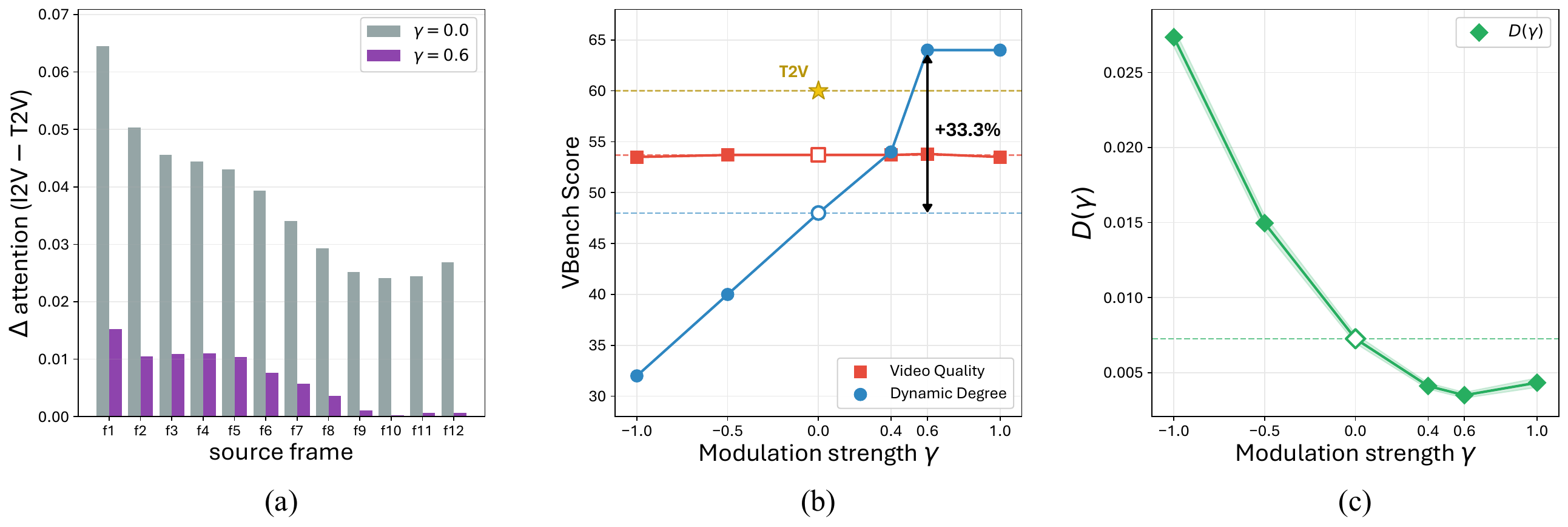}
  \caption{
  \textbf{Modulating reference-frame dominance controls motion dynamics.}
  \textbf{(a)} Absolute difference in reference-frame attention between the vanilla I2V model and the modulated I2V model with $\gamma=0.6$, measured over non-reference query frames.
  \textbf{(b)} Dynamic Degree and Video Quality as $\gamma$ varies. The yellow star denotes the Dynamic Degree of the paired T2V generation.
  \textbf{(c)} T2V--I2V attention distance $D(\gamma)$ measured by Jensen--Shannon divergence.
  }
\label{fig3}
\vspace{-2mm}
\end{figure*}

To test this, we directly intervene on the attention logits during the first 10\% of denoising steps.
At every self-attention layer, we add a scalar bias to the attention logits from non-reference-frame query tokens to reference-frame key tokens before the softmax operation:
\begin{equation}
\label{eq:probe_modulation}
\tilde{\mathcal{L}}[i,j]
=
\mathcal{L}[i,j]
-
\gamma \cdot
\mathbbm{1}[j \in \mathcal{I}_{f_0}]
\cdot
\mathbbm{1}[i \notin \mathcal{I}_{f_0}]
\end{equation}
where $\mathcal{I}_{f_0}$ denotes the set of token indices belonging to the reference frame, and $\gamma$ represents the strength of the intervention. 
A negative value of $\gamma$ increases the logits assigned to reference-frame keys and thereby strengthens reference-frame dominance, whereas a positive $\gamma$ suppresses these logits and weakens the dominance. 
Fig.~\ref{fig3}(a) shows the resulting change in reference-frame attention for $\gamma=0.6$, confirming that the intervention effectively attenuates the intended attention pathway.

We next examine whether this intervention alters generation behavior with VBench~\citep{huang2024vbench} metrics: Dynamic Degree and Video Quality.
As shown in Fig.~\ref{fig3}(b), increasing $\gamma$ consistently increases Dynamic Degree, moving the I2V model from near-static outputs toward more dynamic generation. 
Notably, Video Quality remains stable over a broad range of $\gamma$ values. 
This shows that weakening reference-frame dominance can improve motion dynamics without degrading visual quality. 
Conversely, negative $\gamma$ strengthens reference-frame dominance and results in more static videos.

Finally, we quantify how the modulation affects the frame-to-frame attention structure relative to the paired T2V generation. 
Let $\mathcal{A}^{\mathrm{I2V}}_{\gamma}$ denote the frame-to-frame attention matrix of the I2V model under a modulation strength of $\gamma$. 
We define the T2V--I2V attention distance as
\begin{equation}
\label{eq:t2v_distance}
D(\gamma)
=
\frac{1}{F-1}
\sum_{a=1}^{F-1}
\mathrm{JSD}\!\left(
\mathcal{A}^{\mathrm{I2V}}_{\gamma}[a,:]
\,\|\, 
\mathcal{A}^{\mathrm{T2V}}[a,:]
\right),
\end{equation}
where $\mathrm{JSD}$ denotes the Jensen--Shannon divergence~\citep{lin2002divergence}, and the average is taken over non-reference query frames. 
A smaller $D(\gamma)$ indicates that the I2V attention pattern is closer to that of the paired T2V model. 
Fig.~\ref{fig3}(c) shows an inverse relationship between the modulation strength and the T2V--I2V attention distance.
In particular, $D(\gamma)$ reaches its minimum value around $\gamma=0.6$, where the Dynamic Degree is also closest to that of the paired T2V generation. 
This consistency confirms that attenuating reference-frame dominance moves the I2V model toward a more T2V-like regime in both attention structure and motion dynamics. To further validate the specificity of this intervention, we provide complementary experimental results in Appendix~\ref{app:additional_mechanism}.

\begin{figure*}[t]
\centering
\includegraphics[clip, width=\linewidth]{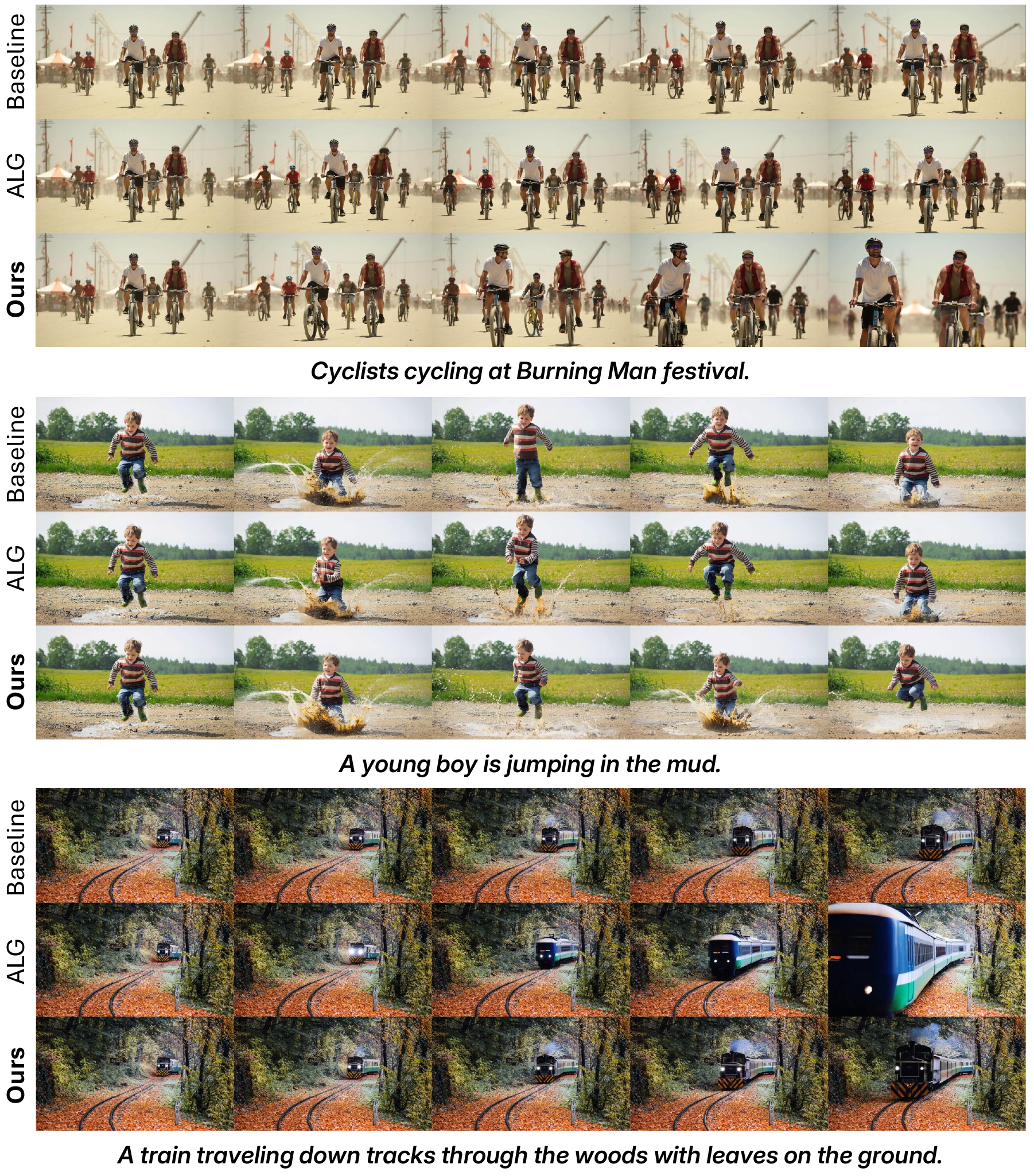}
\caption{
\textbf{Qualitative comparison with vanilla I2V baseline and ALG.} 
The leftmost images are the reference images. 
Our method~(DyMoS) produces substantially more dynamic motion than the vanilla baseline and ALG while preserving fidelity to the reference image. 
In contrast, ALG introduces motion at the cost of visible degradation. We highly encourage readers to visit our project page for a better visualization of our results.
}
\vspace{-3mm}
\label{fig4}
\end{figure*}

\subsection{Proposed method: DyMoS}
\label{sec:DyMoS}

Our analysis provides the key insight that static generation in I2V models can be mitigated by attenuating reference-frame dominance in self-attention. 
Moreover, varying the attenuation strength enables continuous control over motion dynamics.
To this end, we introduce DyMoS~(Dynamic Motion Slider), a simple yet effective inference-time method that subtracts a positive scalar $\gamma$ from the logits associated with reference-frame key tokens by generalizing Eq.~\ref{eq:probe_modulation}:
\begin{equation}
\label{eq:DyMoS_logits}
\tilde{\mathcal{L}}[i,j]
=
\mathcal{L}[i,j]
-
\gamma \cdot \phi(f(i)) \cdot
\mathbbm{1}[j \in \mathcal{I}_{f_0}]
\cdot
\mathbbm{1}[i \notin \mathcal{I}_{f_0}],
\end{equation}
where $f(i)$ denotes the latent-frame index of query token $i$, and $\phi(\cdot)$ is a frame-wise modulation schedule that controls how the modulation strength varies across query frames. 

This design explicitly targets the pathway identified in our analysis: attention from generated frames to reference frame. 
Since the modulation is applied in logit space prior to the softmax operation, DyMoS does not directly discard reference information. 
Instead, it reduces the excessive propagation of reference-frame information to non-reference frames while preserving fidelity to the input image.

DyMoS incorporates two key design choices.
First, we apply the modulation only during the first $\lambda \in [0,1]$ fraction of sampling steps and revert to the original attention computation thereafter. 
Here, $\lambda$ determines the transition point between modulated and standard attention.
This approach is motivated by the observation that reference-frame dominance is most pronounced during early denoising steps, as shown in Fig.~\ref{fig2}(b).
Applying the modulation specifically in this early stage enhances motion, while preserving appearance consistency and temporal coherence in later refinement steps.

Second, under CFG in Eq.~\ref{eq:3}, we apply DyMoS only to the text-conditional branch as
\begin{equation}
\label{eq:cfg}
\tilde{\bm{v}}_{\theta}
(\bm{z}_t,t,\bm{c},\bm{z}_{\mathrm{ref}})
=
\bm{v}_\theta(\bm{z}_t,t,\varnothing,\bm{z}_{\mathrm{ref}})
+
\omega (
\bm{v}_\theta^{\tilde{\mathcal{L}}}
(\bm{z}_t,t,\bm{c},\bm{z}_{\mathrm{ref}})
-
\bm{v}_\theta(\bm{z}_t,t,\varnothing,\bm{z}_{\mathrm{ref}})
),
\end{equation}
where $\bm{v}_\theta^{\tilde{\mathcal{L}}} (\bm{z}_t,t,\bm{c},\bm{z}_{\mathrm{ref}})$ denotes the text-conditional prediction computed with modulated attention logits $\tilde{\mathcal{L}}$, and $\bm{v}_\theta(\bm{z}_t,t,\varnothing,{\bm{z}}_{\mathrm{ref}})$ is the standard prediction under the null text condition.
This design steers the guided update away from reference-frame dominance while keeping the null-text branch unchanged.
The full sampling algorithm for DyMoS is provided in Appendix~\ref{app:algorithm}.


\begin{table*}[t]
\centering
\small
\caption{Quantitative results across various I2V backbones on VBench-I2V test set.}
\vspace{-1mm}
\label{tab:1}
\begin{tabular}{c c cc c c c}
\toprule
\multirow{2}{*}{Model\vspace{-5pt}} 
& \multirow{2}{*}{Method\vspace{-5pt}}
& \multicolumn{2}{c}{Dynamic Degree}
& \multirow{2}{*}{Video Quality\vspace{-5pt}}
& \multirow{2}{*}{ViCLIP\vspace{-5pt}}
& \multirow{2}{*}{\tworow{Vision}{Reward}\vspace{-5pt}} \\
\cmidrule(lr){3-4}
& & VBench & VideoScore & & & \\
\midrule
\multirow{3}{*}{Wan 2.2-14B}
& Baseline & 51.7 & 2.83 & \textbf{58.0} & 0.2614 & 0.143 \\
& ALG & 55.3 & 2.83 & 57.7 & 0.2616 & 0.140 \\
& \cellcolor{green!15}\textbf{Ours} & \cellcolor{green!15}\textbf{64.8} & \cellcolor{green!15}\textbf{2.84} & \cellcolor{green!15}57.7 & \cellcolor{green!15}\textbf{0.2621} & \cellcolor{green!15}\textbf{0.143} \\
\midrule
\multirow{3}{*}{Wan 2.1-14B}
& Baseline & 48.7 & 3.21 & \textbf{57.7} & 0.2601 & \textbf{0.132} \\
& ALG & 52.5 & 3.18 & 57.2 & 0.2615 & 0.127 \\
& \cellcolor{green!15}\textbf{Ours} & \cellcolor{green!15}\textbf{54.5} & \cellcolor{green!15}\textbf{3.25} & \cellcolor{green!15}57.4 & \cellcolor{green!15}\textbf{0.2618} & \cellcolor{green!15}0.126 \\
\midrule
\multirow{3}{*}{HunyuanVideo-1.5}
& Baseline & 48.0 & 2.87 & \textbf{58.2} & 0.2624 & 0.140 \\
& ALG & 46.3 & 2.87 & 58.2 & 0.2629 & 0.141 \\
& \cellcolor{green!15}\textbf{Ours} & \cellcolor{green!15}\textbf{55.3} & \cellcolor{green!15}\textbf{2.88} & \cellcolor{green!15}58.1 & \cellcolor{green!15}\textbf{0.2653} & \cellcolor{green!15}\textbf{0.141} \\
\midrule
\multirow{3}{*}{CogVideoX-5B}
& Baseline & 30.9 & 2.90 & \textbf{57.8} & 0.2612 & \textbf{0.140} \\
& ALG & 52.0 & \textbf{2.98} & 57.1 & 0.2647 & 0.127 \\
& \cellcolor{green!15}\textbf{Ours} & \cellcolor{green!15}\textbf{52.9} & \cellcolor{green!15}2.92 & \cellcolor{green!15}57.1 & \cellcolor{green!15}\textbf{0.2652} & \cellcolor{green!15}0.131 \\
\bottomrule
\end{tabular}
\vspace{-3mm}
\end{table*}

\section{Experiments}
\label{sec:experiments}
\subsection{Experimental setup}
\paragraph{I2V backbones and comparison methods.}
To verify the robustness of our method, we conduct experiments on several state-of-the-art I2V models, including Wan 2.2~\citep{wan2025wan}, Wan 2.1~\citep{wan2025wan}, HunyuanVideo-1.5~\citep{hunyuanvideo2025} and CogVideoX-5B~\citep{yang2025cogvideox}. Wan 2.2 and Wan 2.1 adopt a separate self-attention and cross-attention design, while HunyuanVideo-1.5 and CogVideoX-5B adopt MM-DiT architectures. Evaluating on both designs allows us to verify that our findings and method generalize 
across different architectures. We compare DyMoS against the vanilla I2V backbones~(baseline) and ALG~\citep{choi2025enhancing} using the official default configurations. 
For Wan 2.2, we set $\gamma=0.6$ and $\lambda=0.2$. We select $\gamma$ and $\lambda$ separately for each backbone to account for differences in attention scale and denoising schedule, and keep them fixed across all prompts.
Per-backbone configurations are provided in Appendix~\ref{app:config}.

\paragraph{Evaluation metrics.}
We evaluate the generated videos in various aspects: motion dynamics, video quality, text-video alignment, and human-aligned preference. 
Since DyMoS mainly targets enhancing motion dynamics in I2V generation, we adopt \textit{Dynamic Degree} from both VBench~\citep{huang2024vbench} and VideoScore~\citep{he2024videoscore} as our primary metric. 
For video quality, we report \textit{Video Quality}, an aggregate score computed over selected VBench quality-related dimensions. 
For text-video alignment, we use ViCLIP~\citep{wang2024internvid} to compute the similarity between the input prompt and the generated video. 
Finally, we include VisionReward~\citep{xu2026visionreward}, a VLM-based human-aligned reward model for image and video generation. 
Further details on evaluation metrics are provided in Appendix~\ref{app:metrics}.
\paragraph{Datasets.}
We assess our method on three datasets. 
For our primary evaluation, we use 246 image-caption pairs from the VBench-I2V test set~\citep{huang2024vbench}, which consists of diverse scenes and motion types. 
We additionally evaluate on 200 image-caption pairs sampled from PE Video Dataset (PVD)~\citep{bolya2026perception}, a large-scale video-caption dataset containing both synthetic and human-annotated captions, from which we use the first frame of each video as the conditioning image. 
Finally, we include VidProM~\citep{wang2024vidprom}, a large-scale text-to-video prompt dataset, to assess generalization to diverse text prompts. 
We randomly sample 500 prompts from VidProM and generate corresponding reference images using Flux.1-dev~\citep{flux2024}. 
Further details on each dataset are provided in Appendix~\ref{app:dataset}.

\subsection{Main results}
\paragraph{Qualitative evaluation.}
As illustrated in Fig.~\ref{fig4}, DyMoS achieves a better balance between motion dynamics and reference-image fidelity than both the vanilla I2V baseline and ALG. 
The vanilla I2V baseline often produces nearly static videos, where the generated frames remain overly close to the reference image even when the prompt describes object motion. 
ALG improves motion dynamics over the baseline, but it can introduce visible degradation due to its input-level attenuation. 
For example, in the train sequence, ALG induces motion but also changes the appearance of the train as the video progresses, indicating a loss of reference fidelity. 
In contrast, DyMoS produces substantially more dynamic videos while faithfully preserving the reference image, as shown in the first two examples. 
In particular, in the boy example, DyMoS sustains the jumping motion over more frames, resulting in an additional jump cycle compared to the other methods.
Additional qualitative examples are provided in Appendix~\ref{app:additional_quali}.

\paragraph{Quantitative evaluation.}
Tab.~\ref{tab:1} reports quantitative results across four different I2V backbones on VBench-I2V. 
Our method consistently achieves higher Dynamic Degree than both the baseline and ALG, demonstrating its effectiveness in improving motion dynamics. 
Crucially, DyMoS maintains Video Quality scores comparable to ALG while achieving substantially higher Dynamic Degree. This indicates that our attention-level intervention improves motion dynamics more effectively than input-level attenuation, while preserving visual quality.
Our method also achieves the highest ViCLIP scores among the compared methods, suggesting that reducing reference-frame dominance allows the model to better follow motion described in text prompts. 
In addition, VisionReward scores remain stable or improve, indicating that enhanced motion dynamics do not degrade overall perceptual quality and human preference.
Tab.~\ref{tab:2} further validates the effectiveness of our method across various benchmark datasets, supporting its robust performance across diverse reference images and text prompts. We further provide a comparison of computation costs in Appendix~\ref{app:cost}.

\begin{figure*}[t]
\centering
\includegraphics[clip, width=\linewidth]{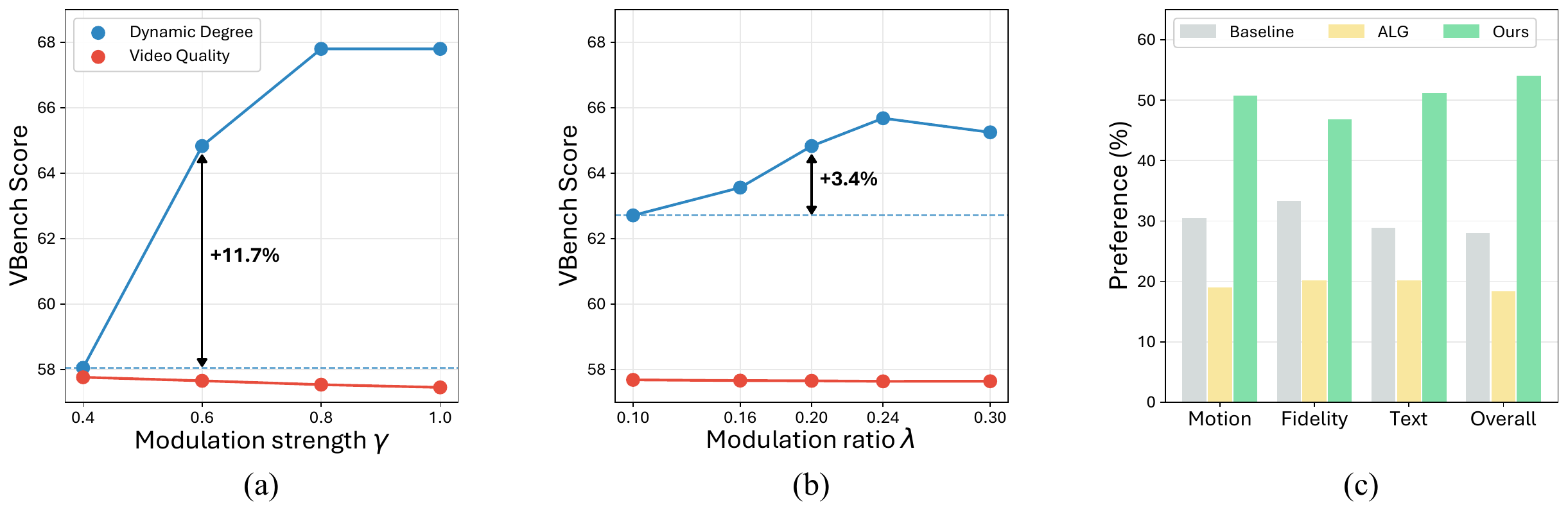}
\vspace{-4mm}
\caption{\textbf{Hyperparameter analysis and user study.} 
(a) Effect of modulation strength $\gamma$.
(b) Effect of modulation step ratio $\lambda$.
(c) User study results. The final preference score for each method is computed as the fraction of times it is selected under each criterion.
}
\vspace{-3mm}
\label{fig:ablation}
\end{figure*}

\paragraph{User study.}
We conduct a user study with 30 participants to complement the automated evaluation. 
Each participant evaluates 25 video triplets, each comprising videos generated from the same reference image and text prompt using the vanilla baseline, ALG, and DyMoS, with method names hidden and presentation order randomized. Participants select the best video according to four criteria: \textit{Motion Dynamicness}, \textit{Reference-Frame Fidelity}, \textit{Text Alignment}, and \textit{Overall Preference}. As shown in Fig.\ref{fig:ablation}(c), DyMoS is preferred most frequently across all four criteria, demonstrating its perceptual advantages. Further details are provided in the Appendix~\ref{app:user_study}.

\begin{table*}[t]
\centering
\small
\caption{Quantitative results with Wan 2.2-14B across various benchmark datasets.}
\vspace{-1mm}
\label{tab:2}
\begin{tabular}{c c cc c c c}
\toprule
\multirow{2}{*}{Dataset\vspace{-5pt}} 
& \multirow{2}{*}{Method\vspace{-5pt}}
& \multicolumn{2}{c}{Dynamic Degree}
& \multirow{2}{*}{Video Quality\vspace{-5pt}}
& \multirow{2}{*}{ViCLIP\vspace{-5pt}}
& \multirow{2}{*}{\tworow{Vision}{Reward}\vspace{-5pt}} \\
\cmidrule(lr){3-4}
& & VBench & VideoScore & & & \\
\midrule
\multirow{3}{*}{VBench-I2V} 
& Baseline & 51.7 & 2.83 & \textbf{58.0} & 0.2614 & 0.143 \\
& ALG & 55.3 & 2.83 & 57.7 & 0.2616 & 0.140 \\
& \cellcolor{green!15}\textbf{Ours} & \cellcolor{green!15}\textbf{64.8} & \cellcolor{green!15}\textbf{2.84} & \cellcolor{green!15}57.7 & \cellcolor{green!15}\textbf{0.2621} & \cellcolor{green!15}\textbf{0.143} \\
\midrule
\multirow{3}{*}{PVD} 
& Baseline & 79.0 & 2.80 & 55.0 & \textbf{0.2147} & \textbf{0.087} \\
& ALG & 77.5 & 2.80 & 55.1 & 0.2141 & 0.084 \\
& \cellcolor{green!15}\textbf{Ours} & \cellcolor{green!15}\textbf{85.0} & \cellcolor{green!15}\textbf{2.80} & \cellcolor{green!15}\textbf{55.1} & \cellcolor{green!15}0.2143 & \cellcolor{green!15}0.086 \\
\midrule
\multirow{3}{*}{VidProM} 
& Baseline & 54.2 & 2.78 & 57.3 & 0.2427 & \textbf{0.064} \\
& ALG & 59.0 & 2.78 & 58.2 & \textbf{0.2433} & 0.062 \\
& \cellcolor{green!15}\textbf{Ours} & \cellcolor{green!15}{\textbf{67.0}} & \cellcolor{green!15}\textbf{2.80} & \cellcolor{green!15}{\textbf{60.1}} & \cellcolor{green!15}0.2432 & \cellcolor{green!15}0.062 \\
\bottomrule
\end{tabular}
\end{table*}

\subsection{Ablation studies}
\label{sec:ablation}

We conduct ablation studies on two key hyperparameters of DyMoS, the modulation strength $\gamma$ and the modulation step ratio $\lambda$, to investigate how the strength and duration of attention modulation affect motion dynamics and video quality.

\paragraph{Effect of modulation strength $\gamma$.}
To assess the impact of the modulation strength $\gamma$, we vary $\gamma \in \{0.4, 0.6, 0.8, 1.0\}$ and measure Dynamic Degree and Video Quality for each setting. 
As shown in Fig.~\ref{fig:ablation}(a), Dynamic Degree consistently increases as $\gamma$ increases, which is consistent with our analysis in \S\ref{sec:gamma_sweep}.
For example, increasing $\gamma$ from $0.4$ to $0.6$ yields an 11.7\% relative gain in Dynamic Degree. 
In contrast, Video Quality remains nearly stable across the same range. 
These results show that $\gamma$ can control the strength of motion dynamics without substantially degrading video quality.

\paragraph{Effect of modulation step ratio $\lambda$.}
Additionally, we investigate how the modulation step ratio $\lambda$
affects motion dynamics and video quality. 
As shown in Fig.~\ref{fig:ablation}(b), increasing $\lambda$ initially improves Dynamic Degree while Video Quality remains stable. 
The improvement saturates beyond a certain point, as increasing $\lambda$ from $0.24$ to $0.30$ reduces Dynamic Degree.
This demonstrates that applying DyMoS to an appropriate number of initial denoising steps is sufficient to enhance motion, while switching back to the original attention computation helps maintain generation quality.

\subsection{Application: Continuous control over motion dynamics}
\label{sec:application}

\begin{figure}[t]
\centering
\includegraphics[width=\linewidth,trim={0 0 0 0},clip,page=1]{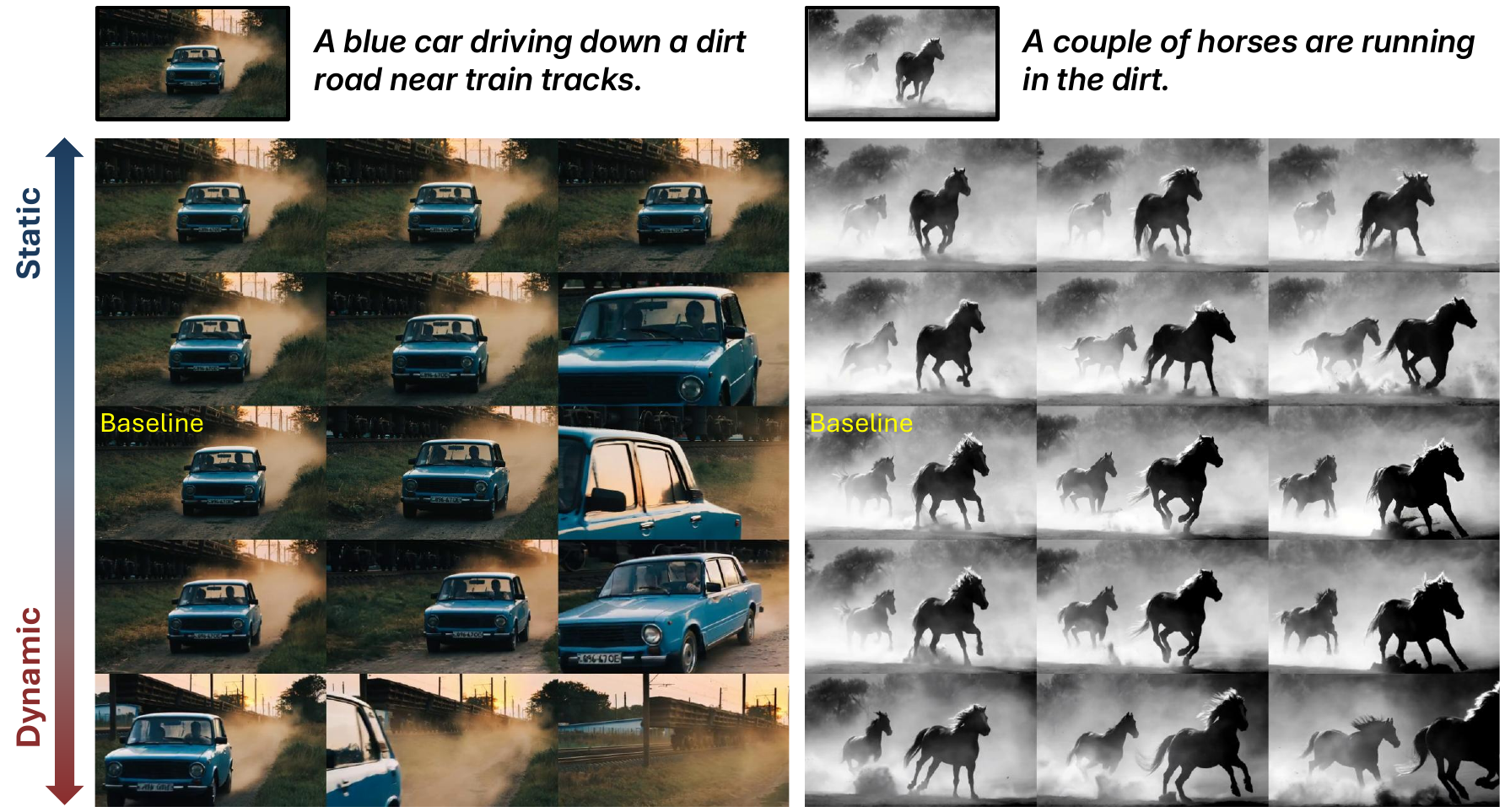}
\vspace{-3mm}
\caption{
\textbf{Continuous control over static-to-dynamic generation with DyMoS.}
Rows from top to bottom correspond to $\gamma \in \{-2,-1,0,0.6,1.0\}$,
with $\gamma=0$ denoting the baseline.
}
\label{fig:static_dynamic}
\vspace{-4mm}
\end{figure}

A key practical advantage of DyMoS is that it provides a simple control knob for continuously adjusting the balance between reference preservation and motion strength. 
Under our convention, $\gamma=0$ recovers the vanilla I2V baseline, $\gamma>0$ weakens reference-frame dominance to encourage stronger motion, and $\gamma<0$ strengthens reference-frame dominance to further suppress motion.

Fig.~\ref{fig:static_dynamic} illustrates this behavior across different values of $\gamma$. At $\gamma=0$, the model strongly preserves the reference image and produces limited motion. 
As $\gamma$ becomes negative, the generated videos become even more static, with the reference image effectively propagated along the temporal axis. 
For example, at $\gamma=-2$, both the car and the horses remain nearly static. 
As $\gamma$ increases above zero, the generated videos exhibit richer motion, such as the car moving along the road with a following camera. 
Moderate positive values of $\gamma$ improve motion while preserving reference fidelity, whereas overly large values suppress reference-frame dominance beyond the level observed in T2V models, yielding much more dynamic motion at the cost of appearance consistency.
These results demonstrate that DyMoS enables continuous control over the static-to-dynamic generation spectrum.

\section{Conclusion}
\label{sec:conclusion}
We present DyMoS~(Dynamic Motion Slider), a training-free and model-agnostic method for improving motion dynamics in image-to-video generation.
Our analysis identifies that static motion bias in I2V models is primarily caused by reference-frame dominance.
DyMoS mitigates this issue by subtracting a log-space bias from reference-frame key logits during initial denoising steps, thereby reducing excessive reference propagation without modifying the input image or model weights. 
With a single parameter $\gamma$, DyMoS offers continuous control over the balance between reference preservation and motion strength. 
Experiments across recent I2V backbones show that DyMoS significantly improves motion dynamics while maintaining reference fidelity and visual quality.

\noindent\textbf{Limitation and future work.} 
Although DyMoS improves motion dynamics across diverse I2V backbones, overly large modulation strengths can reduce appearance consistency and temporal stability by excessively suppressing reference-frame information. In addition, since our method currently uses fixed hyperparameters per backbone, an automatic adapation of the modulation strength to different prompts or motion types remains for future work.


\section*{Acknowledgments}
This work was supported by the Institute of Information \& Communications Technology Planning \& Evaluation (IITP) grants funded by the Korean Government (MSIT) (RS-2020-II201361 and RS-2025-25441838), and by the National Research Foundation of Korea (NRF) grant funded by the Korean government (MSIT) (RS-2024-00338439).

\bibliographystyle{unsrtnat}
\bibliography{reference}

\newpage
\appendix
\section{Full algorithm of DyMoS}
\label{app:algorithm}

\newcommand{\algcmt}[1]{\hfill{\(\triangleright\)~#1}}

\begin{algorithm}[h]
\caption{DyMoS}
\label{alg:method}
\renewcommand{\algorithmicrequire}{\textbf{Input:}}
\renewcommand{\algorithmicensure}{\textbf{Output:}}
\begin{algorithmic}[1]
\REQUIRE Reference image $I_{\mathrm{ref}}$, text prompt $\bm{c}$, total inference steps $N$, guidance scale $\omega$, modulation strength $\gamma$, modulation step ratio $\lambda$, frame-wise schedule $\phi(\cdot)$
\STATE $\bm{z}_{\mathrm{ref}} \leftarrow \mathcal{E}(I_{\mathrm{ref}})$
\STATE $\bm{z} \sim \mathcal{N}(\bm{0}, \bm{I})$
\STATE $\mathcal{I}_{f_0} \leftarrow$ token indices of the reference frame
\FOR{$i = 1$ \textbf{to} $N$}
    \STATE $t \leftarrow \frac{i}{N}$
    \IF{$t < \lambda$}
        \STATE $\tilde{\mathcal{L}}[p,q] \leftarrow
        \mathcal{L}[p,q]
        - \gamma \cdot \phi(f(p)) \cdot
        \mathbbm{1}[q \in \mathcal{I}_{f_0}]
        \cdot
        \mathbbm{1}[p \notin \mathcal{I}_{f_0}]$
        \algcmt{Eq.~\ref{eq:DyMoS_logits}}
    \ELSE
        \STATE $\tilde{\mathcal{L}} \leftarrow \mathcal{L}$
    \ENDIF
    \STATE $\tilde{\bm{v}}_\theta(\bm{z},t,\bm{c},\bm{z}_{\mathrm{ref}})
    \leftarrow
    \bm{v}_\theta(\bm{z},t,\varnothing,\bm{z}_{\mathrm{ref}})
    +
    \omega\Big(
    \bm{v}_\theta^{\tilde{\mathcal{L}}}(\bm{z},t,\bm{c},\bm{z}_{\mathrm{ref}})
    -
    \bm{v}_\theta(\bm{z},t,\varnothing,\bm{z}_{\mathrm{ref}})
    \Big)$
    \algcmt{Eq.~\ref{eq:cfg}}
    \STATE $\bm{z} \leftarrow
    \mathrm{SchedulerStep}(\bm{z},\tilde{\bm{v}}_\theta,t)$
\ENDFOR
\RETURN $\hat{\bm{x}} \leftarrow \mathcal{D}(\bm{z})$
\end{algorithmic}
\end{algorithm}

\section{Additional details on analysis}
\label{app:sa_analysis_details}

\subsection{Experimental setting}
\label{app:sa_analysis_setting}

We provide the experimental procedure used to obtain the frame-to-frame self-attention comparison reported in \S\ref{sec:sa_comparison} and Fig.~\ref{fig2}.

\paragraph{Paired T2V/I2V generation.} 
We construct paired T2V and I2V generations using CogVideoX-5B. 
We first generate a video $\bm{v}^{\text{T2V}}$ with CogVideoX-5B-T2V using a text prompt $\bm{c}$ sampled from T2V-CompBench and a fixed random seed. 
We then extract the first frame $\bm{x}_{\text{ref}}=\bm{v}^{\text{T2V}}[0]$ and feed it together with the same prompt $\bm{c}$ and the same seed to CogVideoX-5B-I2V to obtain $\bm{v}^{\text{I2V}}$. 
We keep the sampling hyperparameters matched between the two runs, including the number of denoising steps, classifier-free guidance scale, resolution, and number of latent frames. 
We repeat the procedure above for $N=50$ prompts sampled from T2V-CompBench, covering a range of motion types, including camera motion, object motion, and human action, as well as diverse scene categories. 
All reported attention statistics are averaged over this prompt set.

\paragraph{Attention extraction.}
At each denoising step $s\in\{0,\dots,N-1\}$, self-attention layer $\ell$,
and attention head $h$, we record the post-softmax attention matrix
$\mathcal{A}^{(s,\ell,h)}\in\mathbb{R}^{L\times L}$, where
$L$ is the total number of latent tokens. We convert this
token-level attention into a frame-to-frame attention matrix by
summing attention over key tokens within each frame and averaging over
query tokens within each query frame:
\begin{equation}
\bar{\mathcal{A}}^{(s,\ell,h)}[a,b]
=
\frac{1}{|\mathcal{I}_a|}
\sum_{i \in \mathcal{I}_a}
\sum_{j \in \mathcal{I}_b}
\mathcal{A}^{(s,\ell,h)}[i,j],
\end{equation}
where $\mathcal{I}_a$ and $\mathcal{I}_b$ denote the token indices
belonging to frames $a$ and $b$, respectively. This quantity represents
the average fraction of attention mass that query tokens in frame $a$
assign to key tokens in frame $b$. For Fig.~\ref{fig2}, we compute $\mathcal{A}^{\mathrm{T2V}}$ and $\mathcal{A}^{\mathrm{I2V}}$ by averaging $\bar{\mathcal{A}}^{(s,\ell,h)}$ over the initial 10\% of denoising steps, attention heads, layers, and prompts.

\subsection{Further validation of reference-frame dominance}
\label{app:additional_mechanism}

We provide further validation of reference-frame dominance in \S\ref{sec:gamma_sweep}.

\paragraph{Ablation study on suppression target.} We examine whether motion enhancement is specific to suppressing reference-frame attention or can also arise from suppressing other attention pathways.
Using CogVideoX-5B and 50 prompts from T2V-CompBench, we compare three suppression targets: reference-frame keys, keys from a randomly selected non-reference frame, and a random token subset containing the same number of keys as the reference frame.
We use the same modulation strength and denoising interval across the three interventions.
As shown in Tab.~\ref{tab:additional_suppression_targets}, suppressing reference-frame keys increases Dynamic Degree from 48.0 to 64.0, while the two control interventions yield scores of 50.0 and 48.0.
Video Quality remains comparable to the baseline under reference-frame suppression.
Together with the bidirectional modulation results in the main paper, these controls support a specific contribution of the reference-frame pathway to motion suppression.
They do not establish that reference-frame dominance is the sole source of static generation.

\begin{table*}[h]
\centering
\caption{\textbf{Experiments on varying suppression target.}}
\vspace{-1mm}
\label{tab:additional_suppression_targets}
\begin{tabular}{lrr}
\toprule
Suppression target & DD $\uparrow$ & VQ $\uparrow$ \\
\midrule
None (baseline) & 48.0 & 53.7 \\
Random non-reference frame & 50.0 & 53.2 \\
Random token subset & 48.0 & 53.3 \\
Reference frame (DyMoS) & 64.0 & 53.8 \\
\bottomrule
\end{tabular}
\end{table*}

\paragraph{Generalization across backbones.}
We extend the paired T2V/I2V analysis to Wan~2.2, Wan~2.1, and HunyuanVideo-1.5 using 50 prompts from T2V-CompBench.
Following the procedure in Appendix~\ref{app:sa_analysis_setting}, we measure the attention mass assigned by each non-reference query frame to reference-frame keys and average over the first 10\% of denoising steps.
Tab.~\ref{tab:additional_backbone_attention} reports the relative increase in this mass for I2V compared with the corresponding T2V model at selected latent frames.
All three backbones exhibit increased reference attention across multiple non-reference frames.
The pattern varies across architectures and temporal positions.
In particular, the final reported frame has a negative difference for Wan~2.2 and Wan~2.1.
These results extend the evidence beyond CogVideoX-5B without implying a uniform increase at every frame.

\begin{table*}[h]
\centering
\small
\setlength{\tabcolsep}{5pt}
\caption{\textbf{Reference attention across additional backbones.}
Relative change in reference-frame attention mass from T2V to I2V, in percent.
The reference frame is $f_0$.}
\vspace{-1mm}
\label{tab:additional_backbone_attention}
\begin{tabular}{lccccccccc}
\toprule
Backbone
& $f_1$ & $f_2$ & $f_3$ & $f_4$ & $f_5$
& $\cdots$ & $f_{18}$ & $f_{19}$ & $f_{20}$ \\
\midrule
Wan~2.2
& $+35.9$ & $+53.3$ & $+31.2$ & $+45.9$ & $+35.9$
& $\cdots$ & $+25.3$ & $+11.1$ & $-13.1$ \\

Wan~2.1
& $+41.5$ & $+45.6$ & $+28.2$ & $+27.7$ & $+16.6$
& $\cdots$ & $+14.6$ & $+0.7$ & $-4.7$ \\

HunyuanVideo-1.5
& $+55.9$ & $+61.0$ & $+79.0$ & $+74.7$ & $+81.3$
& $\cdots$ & $+151.5$ & $+146.9$ & $+114.8$ \\
\bottomrule
\end{tabular}
\end{table*}

\paragraph{Evolution over denoising steps.}
We evaluate the T2V--I2V attention distance in Eq.~\ref{eq:t2v_distance} at five denoising steps using CogVideoX-5B.
For this analysis, attention statistics are aggregated separately at each reported step.
As shown in Tab.~\ref{tab:additional_step_attention}, the distance is largest at step~5 among the measured steps and decreases to 0.0012 at the final step.
This indicates that the frame-to-frame attention distributions differ more strongly during early denoising than during later refinement.

\begin{table*}[h]
\centering
\caption{\textbf{T2V--I2V Attention distance over denoising steps.} Step~50 is the final sampling step.}
\vspace{-1mm}
\label{tab:additional_step_attention}
\begin{tabular}{lccccc}
\toprule
Denoising step & 1 & 5 & 10 & 25 & 50 \\
\midrule
Attention distance & 0.0079 & 0.0109 & 0.0082 & 0.0030 & 0.0012 \\
\bottomrule
\end{tabular}
\end{table*}

\section{Additional details on the experimental setup}
\label{app:impl}

\begin{table}[t]
\centering
\small
\caption{Detailed configurations for each I2V backbone.}
\label{tab4}
\vspace{2mm}
\setlength{\tabcolsep}{6pt}
\begin{tabular}{l l cccc}
\toprule
& Models & Wan 2.2-14B & Wan 2.1-14B & HunyuanVideo-1.5 & CogVideoX-5B \\
\midrule
\multirow{5}{*}{Base}
& Resolution       & 81$\times$832$\times$480 & 81$\times$832$\times$480 & 81$\times$832$\times$480 & 49$\times$720$\times$480 \\
& FPS              & 16 & 16 & 16 & 8 \\
& Denoising steps              & 40 & 40 & 50 & 50 \\
& CFG scale        & 3.5 & 5.0 & 6.0 & 6.0 \\
& Scheduler        & UniPC & UniPC & FlowMatch-Euler & DDIM \\
\midrule
\multirow{3}{*}{DyMoS}
& $\gamma$          & 0.6 & 1.0 & 1.0 & 0.8 \\
& $\lambda$         & 0.2 & 0.25 & 0.2 & 0.24 \\
& $\phi(\cdot)$     & 1 & 1 & 1 & 1 \\
\bottomrule
\end{tabular}
\end{table}
\begin{table}[t]
\centering
\small
\caption{Model checkpoints used in our main experiments.}
\vspace{2mm}
\label{tab:model_checkpoints}
\begin{tabular}{l l l}
\toprule
\textbf{Model} & \textbf{Type} & \textbf{Source} \\
\midrule
Wan~2.2
& I2V & \raggedright\arraybackslash \url{https://huggingface.co/Wan-AI/Wan2.2-I2V-A14B-Diffusers} \\
Wan~2.1
& I2V & \raggedright\arraybackslash \url{https://huggingface.co/Wan-AI/Wan2.1-I2V-14B-480P-Diffusers} \\
HunyuanVideo-1.5
& I2V & \raggedright\arraybackslash \url{https://huggingface.co/tencent/HunyuanVideo-1.5} \\
CogVideoX-5B
& I2V & \raggedright\arraybackslash \url{https://huggingface.co/zai-org/CogVideoX-5b-I2V} \\
\bottomrule
\end{tabular}
\end{table}
\subsection{Implementation and setup}
\label{app:config}
\paragraph{Implementation details.}
We implement DyMoS on top of the official open-source releases of Wan~2.2, Wan~2.1, HunyuanVideo-1.5 and CogVideoX-5B. The model checkpoints we used in our experiments are provided in Tab.~\ref{tab4}. DyMoS is applied identically across all backbones. During the first $\lambda N$ denoising steps, we subtract a bias $\gamma$ to the pre-softmax self-attention logits for query-key pairs where the key is the frame-0 video token and the query is a non-frame-0 video token. The frame-0$\to$frame-0 attention block is exempted to preserve the reference frame. 

\paragraph{Configuration.}
Tab.~\ref{tab4} summarizes the per-backbone sampling configurations used in all experiments, including resolution, FPS, denoising steps, CFG scale, scheduler, and our hyperparameters $(\gamma, \lambda, \phi_f)$. 
All experiments are conducted on either a single NVIDIA H100 GPU or a single NVIDIA A100 GPU. 
For each backbone, all methods are run on the same GPU type, and no multi-GPU inference is used. 
Random seeds are fixed across methods within each pair to ensure that comparisons better isolate the effect of the method itself. 
For ALG~\citep{choi2025enhancing}, we use the official implementation~\footnote{https://github.com/choi403/ALG} and the recommended hyperparameters reported in the paper.

\subsection{Evaluation metrics}
\label{app:metrics}

We provide details of the evaluation metrics used throughout our
experiments, including VBench~\citep{huang2024vbench},
VideoScore~\citep{he2024videoscore}, ViCLIP~\citep{wang2024internvid},
and VisionReward~\citep{xu2026visionreward}. Our evaluation covers four
aspects: motion dynamics, video quality, text-video alignment, and
human-aligned preference.

\paragraph{Motion dynamics.}
We evaluate motion dynamics using \textit{Dynamic Degree} from both
VBench and VideoScore. VBench Dynamic Degree is designed to measure whether a generated video contains sufficiently large motion by computing the highest 5\% optical flows across frames using RAFT~\citep{teed2020raft}. VideoScore Dynamic Degree is predicted by a VLM-based video evaluator trained on human feedback. We report both metrics to verify that DyMoS improves motion dynamics under both a conventional metric and a learned human-aligned evaluator.

\paragraph{Video quality.}
We report \textit{Video Quality} as an aggregate score over selected
VBench quality-related dimensions as done in previous works~\citep{yang2025cogvideox, jeong2025track4gen, ge2025flashi2v, choi2025enhancing, lee2026memory}. Specifically, we compute
\[
\mathrm{VideoQuality}
=
0.1 \cdot \mathrm{TF}
+0.25 \cdot \mathrm{MS}
+0.1 \cdot \mathrm{AQ}
+0.25 \cdot \mathrm{IQ},
\]
where TF, MS, AQ, and IQ denote Temporal Flickering, Motion Smoothness,
Aesthetic Quality, and Imaging Quality, respectively. These dimensions
capture complementary aspects of perceptual video quality: TF penalizes
frame-to-frame flickering artifacts, MS measures the temporal coherence
of motion, and AQ/IQ assess the visual appeal and low-level imaging
quality of individual frames.
The weights follow the official VBench aggregation; we omit Dynamic
Degree, Subject Consistency, and Background Consistency to isolate video
quality from motion strength and reference preservation.

\paragraph{Text-video alignment.}
We use ViCLIP to evaluate text-video semantic alignment. ViCLIP is a
ViT-L-based video-text representation model trained on InternVid with
contrastive video-language learning, mapping videos and text prompts
into a shared embedding space. We compute the similarity between the
input prompt and the generated video representation as a global measure
of text-video semantic matching.

\paragraph{Human-aligned preference.}
We additionally evaluate generated videos with VisionReward, a
VLM-based reward model for image and video generation. VisionReward
decomposes human visual preferences into multiple fine-grained
assessment questions (e.g. prompt alignment, fidelity, motion realism, stability) and combines the corresponding judgments through interpretable linear weighting to produce a human-aligned reward score.
It is trained to capture multi-dimensional human preferences over both
images and videos, including dynamic aspects that are important for
video evaluation. We use VisionReward to complement benchmark-specific
metrics and assess whether DyMoS's outputs are also favored by a
learned human-aligned evaluator.

\subsection{Evaluation datasets}
\label{app:dataset}

We evaluate DyMoS on three datasets covering complementary scenarios: a standard I2V benchmark, a recently released video-caption dataset, and a large-scale prompt collection paired with synthesized images. Note that we adopt the same dataset suite as ALG to ensure consistent comparison.

\paragraph{VBench-I2V.}
\textit{VBench-I2V} is the official I2V benchmark of VBench, consisting of image-caption pairs for evaluating image-to-video generation. Since our study focuses on video motion dynamics, we use the 246 image-caption pairs obtained after excluding subsets designed for background-quality evaluation and camera-motion instruction following. The provided reference images are used directly as conditioning inputs. We do not report the I2V-specific subject/background consistency dimensions, as our evaluation focuses on motion and video quality.

\paragraph{PE Video Dataset (PVD).}
\textit{PE Video Dataset (PVD)} is a 
large-scale video-caption dataset released alongside the Perception 
Encoder, containing both synthetic and human-annotated video captions. 
We randomly sample 200 videos from PVD and use the first frame of 
each video as the conditioning image, paired with the corresponding 
caption as the text prompt. PVD allows us to evaluate DyMoS on 
real-world videos with naturally diverse motion patterns.

\paragraph{VidProM.}
\textit{VidProM} is a large-scale text-to-video 
prompt dataset originally collected to study real user prompts in T2V 
generation. We randomly sample 500 prompts from VidProM and generate 
the corresponding conditioning images using a text-to-image model
Flux.1-dev. This setup probes whether DyMoS 
generalizes to diverse text prompts beyond curated I2V benchmarks.

\subsection{User study protocol}
\label{app:user_study}

To further evaluate beyond automated metrics, we conduct a user study via
MTurk~\citep{crowston2012amazon} following the previous works~\citep{shin2025video, Shin2024bw, park2024kinetic, lee2026universal, jeong2025reangle}. 
For each trial, participants are shown three videos generated from the same reference image and text prompt by the vanilla I2V baseline, ALG, and DyMoS.
Method names are hidden, and the video order is randomized.

Participants answer four questions:
\begin{enumerate}[leftmargin=1.5em,itemsep=2pt,topsep=2pt]
    \item \textbf{Motion:} Which video has the most dynamic and realistic
    motion? Examples include water ripples, cloth movement, human action,
    and camera motion.
    \item \textbf{Fidelity:} Which video best preserves the appearance of
    the reference image throughout the sequence? Examples include the
    subject, background, and colors. 
    \item \textbf{Text alignment:} Which video most faithfully reflects
    the content described in the text prompt?
    \item \textbf{Overall preference:} Overall, which video do you prefer?
\end{enumerate}

We collect 30 responses for each question over 25 randomly sampled image--prompt pairs from VBench-I2V. 
This yields 750 participant--triplet evaluations and 3,000 criterion-level judgments.
The final preference score for each method is computed as the fraction of times it is selected under each criterion.

\section{Additional experimental results}

\subsection{{Full quantitative results}}
\label{app:additional_fidelity}


We report the full quantitative results for all four backbones evaluated on the VBench-I2V dataset in Tab.~\ref{tab:full_result}. Specifically, we report each scores of VBench~\citep{huang2024vbench} including Subject Consistency (SC), Background Consistency (BC), Temporal Flickering (TF), Motion Smoothness (MS), Aesthetic Quality (AQ) and Imaging Quality (IQ) in addition to the aggregate score. Since SC and BC can decrease as the generated motion becomes more dynamic, we additionally assess reference fidelity using per-frame CLIP-I similarity score. Despite producing stronger motion, DyMoS achieves comparable or higher CLIP-I scores than ALG across all backbones, suggesting a better balance between motion dynamics and reference fidelity. 


\begin{table*}[t]
\centering
\small
\setlength{\tabcolsep}{3pt}
\caption{Full quantitative results across various I2V backbones on the VBench-I2V test set.}
\vspace{-1mm}
\label{tab:full_result}
\resizebox{\textwidth}{!}{%
\begin{tabular}{llcc|ccccccc|ccc}
\toprule
\multirow{2}{*}{Backbone}
& \multirow{2}{*}{Method}
& \multicolumn{2}{c|}{Dynamic Degree}
& \multicolumn{7}{c|}{VBench}
& \multirow{2}{*}{CLIP-I}
& \multirow{2}{*}{ViCLIP}
& \multirow{2}{*}{\tworow{Vision}{Reward}} \\
\cmidrule(lr){3-4}
\cmidrule(lr){5-11}
& & VBench & VideoScore
& Video Quality & SC & BC & TF & MS & AQ & IQ
& & & \\
\midrule

\multirow{3}{*}{Wan 2.2-14B}
& Baseline
& 51.7 & 2.83
& \textbf{58.0} & \textbf{94.7} & \textbf{94.9} & \textbf{96.4} & \textbf{98.0} & \textbf{64.3} & \textbf{69.6}
& \textbf{94.4} & 0.2614 & 0.143 \\
& ALG
& 55.3 & 2.83
& 57.7 & 94.1 & 94.5 & 95.9 & 97.7 & 63.8 & 69.6
& 93.9 & 0.2616 & 0.140 \\
& \cellcolor{green!15}\textbf{DyMoS}
& \cellcolor{green!15}\textbf{64.8}
& \cellcolor{green!15}\textbf{2.84}
& \cellcolor{green!15}57.7
& \cellcolor{green!15}93.5
& \cellcolor{green!15}94.1
& \cellcolor{green!15}95.9
& \cellcolor{green!15}97.7
& \cellcolor{green!15}63.8
& \cellcolor{green!15}69.1
& \cellcolor{green!15}94.2
& \cellcolor{green!15}\textbf{0.2621}
& \cellcolor{green!15}\textbf{0.143} \\
\midrule

\multirow{3}{*}{Wan 2.1-14B}
& Baseline
& 48.7 & 3.21
& \textbf{57.7} & \textbf{94.5} & \textbf{95.6} & \textbf{97.0} & \textbf{98.2} & \textbf{60.9} & \textbf{69.4} & \textbf{90.6} & 0.2601 & \textbf{0.132} \\
& ALG
& 52.5 & 3.18
& 57.2 & 92.6 & 94.9 & 96.8 & 98.1 & 60.5 & 67.7
& 89.3 & 0.2615 & 0.127 \\
& \cellcolor{green!15}\textbf{DyMoS}
& \cellcolor{green!15}\textbf{54.5}
& \cellcolor{green!15}\textbf{3.25}
& \cellcolor{green!15}57.4
& \cellcolor{green!15}93.2
& \cellcolor{green!15}94.9
& \cellcolor{green!15}96.5
& \cellcolor{green!15}98.0
& \cellcolor{green!15}60.5
& \cellcolor{green!15}68.9
& \cellcolor{green!15}89.4
& \cellcolor{green!15}\textbf{0.2618}
& \cellcolor{green!15}0.126 \\
\midrule

\multirow{3}{*}{HunyuanVideo-1.5}
& Baseline
& 48.0 & 2.87
& \textbf{58.2} & \textbf{95.0} & \textbf{94.9} & \textbf{97.7} & \textbf{99.1} & \textbf{61.2} & \textbf{70.2}
& \textbf{90.5} & 0.2624 & 0.140 \\
& ALG
& 46.3 & 2.87
& 58.2 & 94.7 & 94.8 & 97.6 & 99.1 & 61.2 & 70.0
& 89.9 & 0.2629 & 0.141 \\
& \cellcolor{green!15}\textbf{DyMoS}
& \cellcolor{green!15}\textbf{55.3}
& \cellcolor{green!15}\textbf{2.88}
& \cellcolor{green!15}58.1
& \cellcolor{green!15}93.1
& \cellcolor{green!15}94.1
& \cellcolor{green!15}97.6
& \cellcolor{green!15}99.1
& \cellcolor{green!15}61.0
& \cellcolor{green!15}70.0
& \cellcolor{green!15}89.9
& \cellcolor{green!15}\textbf{0.2653}
& \cellcolor{green!15}\textbf{0.141} \\
\midrule

\multirow{3}{*}{CogVideoX-5B}
& Baseline
& 30.9 & 2.90
& \textbf{57.8} & \textbf{95.3} & \textbf{95.8} & \textbf{97.2} & \textbf{98.6} & \textbf{60.5} & \textbf{69.6}
& 91.6 & 0.2612 & \textbf{0.140} \\
& ALG
& 52.0 & \textbf{2.98}
& 57.1 & 92.0 & 94.0 & 96.3 & 98.0 & 59.2 & 68.0
& 88.6 & 0.2647 & 0.127 \\
& \cellcolor{green!15}\textbf{DyMoS}
& \cellcolor{green!15}\textbf{52.9}
& \cellcolor{green!15}2.92
& \cellcolor{green!15}57.1
& \cellcolor{green!15}88.3
& \cellcolor{green!15}92.4
& \cellcolor{green!15}96.7
& \cellcolor{green!15}98.0
& \cellcolor{green!15}59.2
& \cellcolor{green!15}68.0
& \cellcolor{green!15}89.4
& \cellcolor{green!15}\textbf{0.2652}
& \cellcolor{green!15}0.131 \\
\bottomrule
\end{tabular}%
}
\vspace{-3mm}
\end{table*}

\subsection{Computational cost}
\label{app:cost}
Both ALG and DyMoS are training-free methods, but their runtime overhead comes from different sources. During its active interval, ALG requires an additional diffusion transformer forward pass for low-pass conditioned branch. In contrast, DyMoS keeps the original sampling loop unchanged. We use the official denoising steps in the official configuration of each backbone and the standard batched unconditional/conditional forward pass for classifier-free guidance.

The only extra computation in DyMoS is the self-attention score modification applied during the first $\lambda_{\text{Ours}}N$ denoising steps. This modification is fused into the attention kernel via PyTorch FlexAttention, and therefore adds only a lightweight element-wise operation on top of the existing attention computation, without any additional model forward pass.

Tab.~\ref{tab:timing} reports per-video inference time measured on a single NVIDIA A100 80GB GPU. Although DyMoS uses a larger active interval than ALG for most backbones, its amortized runtime overhead remains between $1.2\%$ and $4.3\%$. By contrast, ALG incurs a larger overhead due to its additional forward pass, reaching up to $10.3\%$ on Wan~2.1.

\vspace{-1mm}
\begin{table}[h]
\centering
\caption{Computational cost comparison of our method with ALG across I2V backbones.}
\vspace{2mm}
\label{tab:timing}
\setlength{\tabcolsep}{6pt}
\begin{tabular}{lcccc}
\toprule
                          & Wan~2.2 & Wan~2.1 & HV-1.5 & CogVideoX \\
\midrule
$\lambda_{\text{ALG}}$    & 0.10  & 0.20  & 0.04  & 0.04  \\
$\lambda_{\text{Ours}}$   & 0.20  & 0.25  & 0.20  & 0.24  \\
\midrule
Baseline (s)              & 831   & 735   & 486   & 232   \\
+ALG (s)                  & 870 (+4.7\%) & 811 (+10.3\%) & 515 (+5.9\%) & 235 (+1.3\%) \\
+Ours (s)                 & 841 (+1.2\%) & 749 (+1.9\%)  & 503 (+3.5\%) & 242 (+4.3\%) \\
\bottomrule
\end{tabular}
\end{table}
\vspace{-1mm}



\subsection{Ablation study on frame-wise modulation schedule}
\label{app:ablation_frame_schedule}

We further analyze the frame-wise modulation schedule $\phi(\cdot)$,
which controls how the modulation strength varies across query frames in
Eq.~\ref{eq:DyMoS_logits}. Let $f(i)\in\{0,\ldots,F-1\}$ denote
the latent-frame index of query token $i$. We compare three simple
schedules:
\begin{align}
\text{Uniform:} \quad
& \phi(f(i)) = 1, \\
\text{Linear:} \quad
& \phi(f(i)) = \frac{f(i)}{F-1}, \\
\text{Log:} \quad
& \phi(f(i)) = \frac{\log(1+f(i))}{\log(F)}.
\end{align}
The Uniform schedule applies the same modulation strength to all
non-reference query frames. The Linear and Log schedules apply weaker
modulation to frames closer to the reference frame and stronger
modulation to later frames, with the Log schedule increasing more rapidly
for early frames and saturating for later ones.

\begin{table*}[h]
\centering
\caption{Ablation study on frame-wise modulation schedule $\phi_f$ on Wan~2.2-14B.}
\label{tab:frame_schedule_ablation}
\begin{tabular}{c cc c c c}
\toprule
\multirow{2}{*}{$\phi_f$\vspace{-5pt}}
& \multicolumn{2}{c}{Dynamic Degree}
& \multirow{2}{*}{Video Quality\vspace{-5pt}}
& \multirow{2}{*}{ViCLIP\vspace{-5pt}}
& \multirow{2}{*}{\tworow{Vision}{Reward}\vspace{-5pt}} \\
\cmidrule(lr){2-3}
& VBench & VideoScore & & & \\
\midrule
Uniform & 64.8 & 2.84 & 57.7 & 0.2621 & 0.143 \\
Linear  & 56.1 & 2.82 & 57.9 & 0.2598 & 0.143 \\
Log     & 60.2 & 2.80 & 57.7 & 0.2597 & 0.141 \\
\bottomrule
\end{tabular}
\end{table*}

Tab.~\ref{tab:frame_schedule_ablation} reports the ablation results on
Wan~2.2 under the same $\gamma$ and $\lambda$ settings as the main
experiments. Among the evaluated schedules, the Uniform schedule achieves
the highest Dynamic Degree while maintaining comparable Video Quality and
text-video alignment. We therefore use the Uniform schedule as the
default setting in the main experiments.

\subsection{Motion evaluation across temporal segments}
Early denoising steps and early video frames represent different axes of generation.
Although DyMoS operates only during early denoising steps, its uniform frame schedule modulates queries from all non-reference frames.
To examine whether the resulting motion enhancement is localized to specific temporal regions, we divide each generated video into four consecutive segments and measure the average top-5\% optical-flow magnitude within each segment.
As shown in Tab.~\ref{tab:additional_temporal_segments}, DyMoS consistently increases motion magnitude across all four temporal segments on Wan~2.2-14B and VBench-I2V.
This indicates that the motion enhancement induced during early denoising steps persists throughout the generated video rather than being concentrated in the early frames.

\begin{table*}[h]
\centering
\caption{{\textbf{Motion magnitude across temporal segments.} Average top-5\% optical-flow magnitude on Wan~2.2-14B and VBench-I2V. Relative gains of DyMoS over the baseline are shown in parentheses.}}
\label{tab:additional_temporal_segments}
\begin{tabular}{lcccc}
\toprule
Method & 0--25\% & 25--50\% & 50--75\% & 75--100\% \\
\midrule
Baseline
& 12.74 & 13.78 & 14.55 & 15.77 \\
DyMoS
& 16.30~(+28.0\%)
& 18.15~(+31.7\%)
& 18.90~(+29.9\%)
& 20.42~(+29.4\%) \\
\bottomrule
\end{tabular}
\end{table*}

\clearpage




\clearpage
\subsection{Additional qualitative results}
\label{app:additional_quali}
We provide additional qualitative comparisons and continuous-control examples. 

\begin{figure}[h]
\centering
\includegraphics[width=\linewidth,trim={0 0 0 0},clip,page=2]{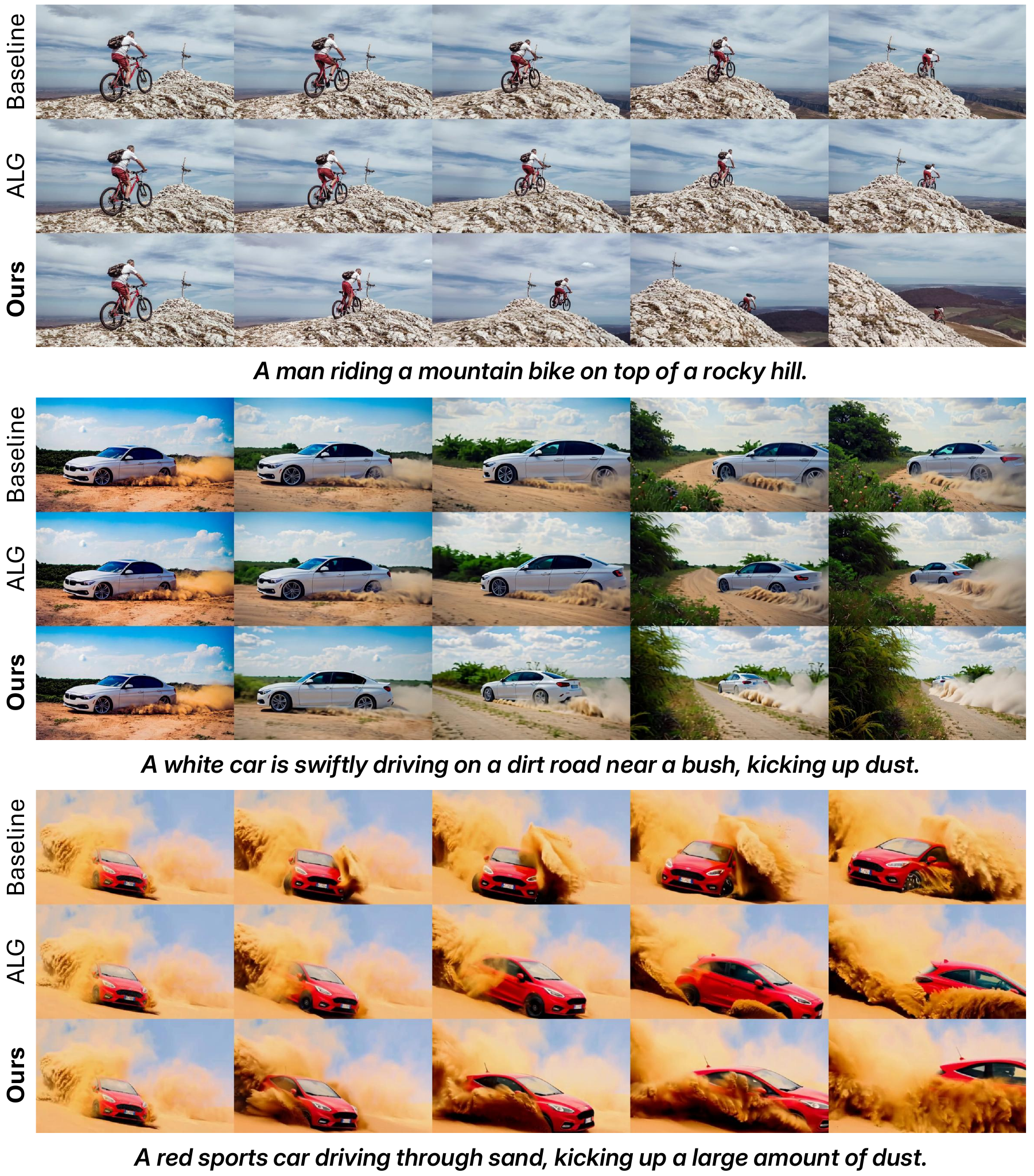}
\caption{\textbf{Additional comparison results with the vanilla baseline and ALG.} The leftmost
images are the reference images. Our method qualitatively outperforms the vanilla baseline and ALG across various cases, demonstrating superior motion dynamics and visual fidelity. (Top) The vanilla baseline and ALG exhibit static scenes of a man riding a mountain bike. In contrast, our method generates fluid and natural riding movements. (Middle and Bottom) Moreover, our method successfully captures complex dynamic interactions, such as cars kicking up dust, while maintaining high visual quality.}
\label{fig6}
\end{figure}

\begin{figure}[t]
\centering
\includegraphics[width=\linewidth,trim={0 0 0 0},clip,page=3]{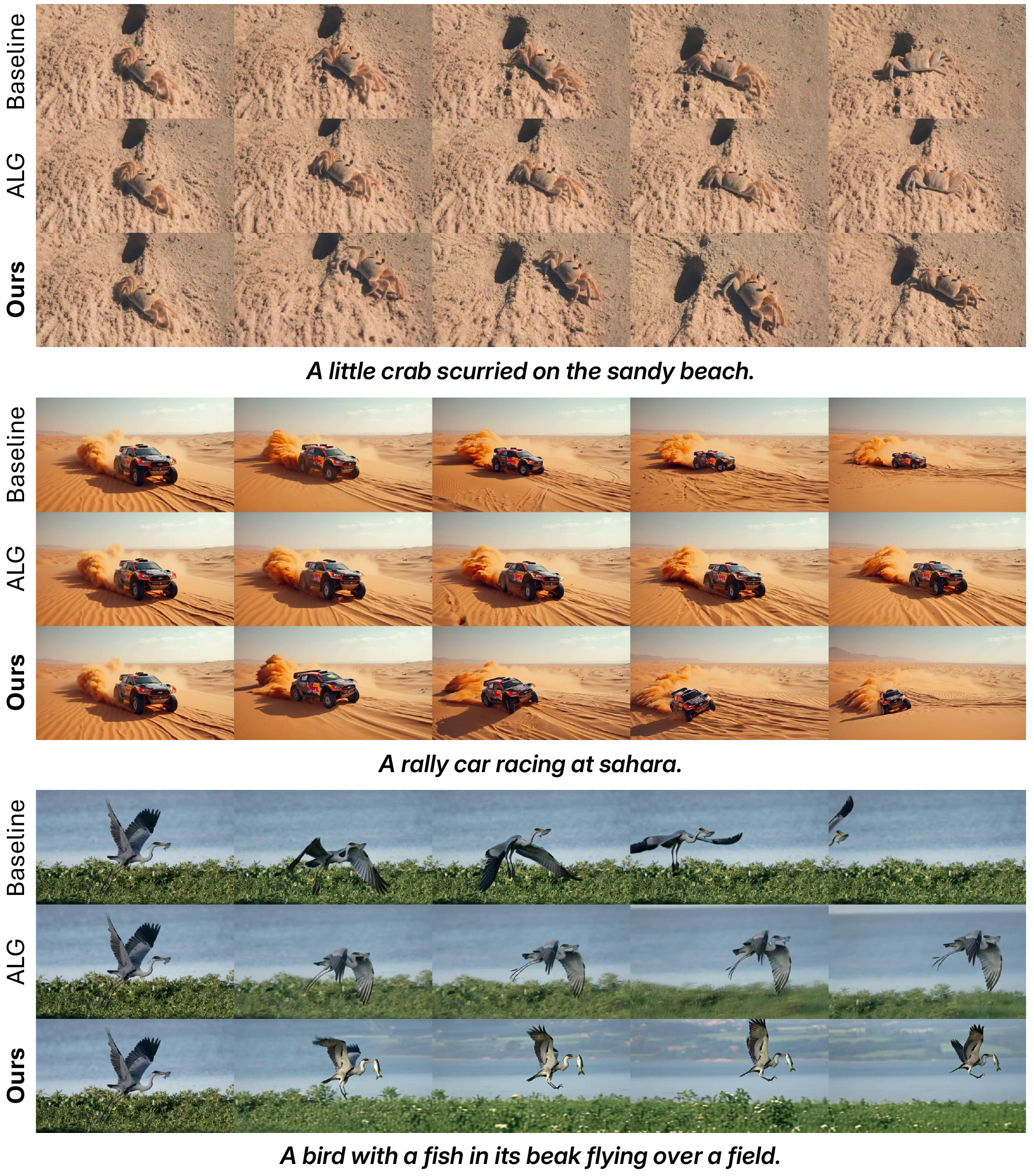}
\caption{\textbf{Additional comparison results with the vanilla baseline and ALG.} The leftmost images are the reference images. (Top) ALG produces highly static scenes where the crab barely moves. In contrast, our method successfully synthesizes vivid and realistic motions. (Bottom) The vanilla baseline produces physically weird motions, such as the bird flying backwards. While ALG generates physically plausible movements, it fails to maintain visual fidelity, exhibiting noticeable distortions in the background.}
\label{fig7}
\end{figure}

\begin{figure}[t]
\centering
\includegraphics[width=\linewidth,trim={0 0 0 0},clip,page=4]{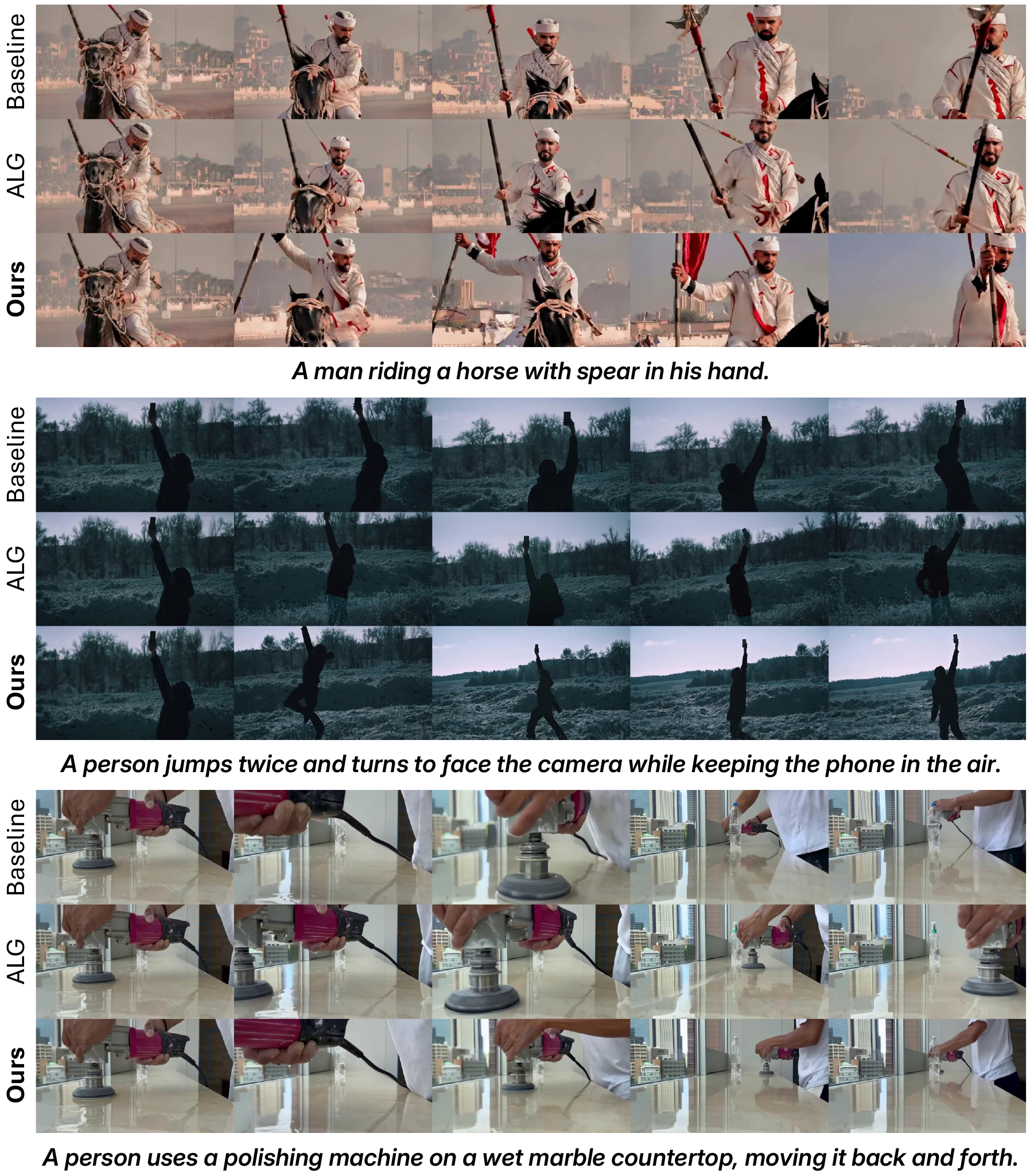}
\caption{\textbf{Additional comparison results with the vanilla baseline and ALG.} The leftmost
images are the reference images. (Top) Both the vanilla baseline and ALG struggle to synthesize dynamic motions, resulting in rigid and unnatural movements of the man. In contrast, our method generates natural motions of the man hoisting a spear. (Middle) Unlike the other methods, DyMoS successfully generates a semantically aligned video in which the person jumps twice. (Bottom) The vanilla baseline produces a physically implausible video where the person's arms penetrate the water bottle, whereas ALG simply fails to generate dynamic motions.}
\label{fig8}
\end{figure}

\begin{figure}[t]
\centering
\includegraphics[width=\linewidth]{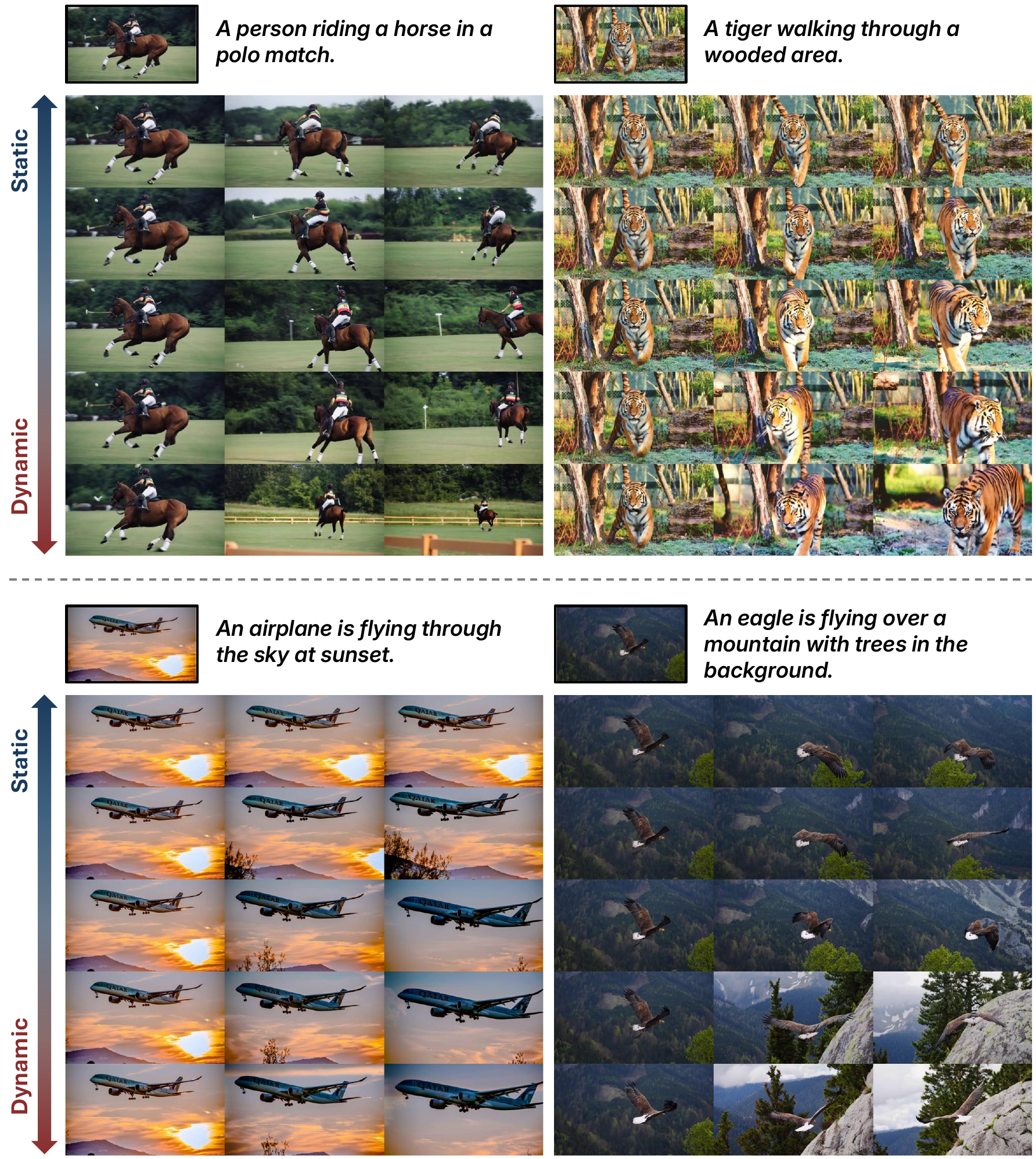}
\caption{
\textbf{Additional examples of continuous control over motion dynamics with DyMoS.} Rows from top to bottom correspond to $\gamma \in \{-2,-1,0,0.6,1.0\}$, with $\gamma=0$ denoting the baseline. (Top Left) The movement of the person riding a horse transitions naturally across the static-to-dynamic spectrum as the parameter increases. (Top Right) Our guidance smoothly scales the tiger's walking speed from slow to fast. Similarly, our method allows for dynamic control over the speeds of the airplane (Bottom Left) and the eagle (Bottom Right).
}
\label{fig:static_dynamic_appendix}
\end{figure}

\clearpage

\clearpage
\section*{NeurIPS Paper Checklist}

\begin{enumerate}

\item {\bf Claims}
    \item[] Question: Do the main claims made in the abstract and introduction accurately reflect the paper's contributions and scope?
    \item[] Answer: \answerYes{} 
    \item[] Justification: The abstract and introduction clearly state the scope of DyMoS, and we detail all claims throughout the main paper and the Appendix. 
    \item[] Guidelines:
    \begin{itemize}
        \item The answer \answerNA{} means that the abstract and introduction do not include the claims made in the paper.
        \item The abstract and/or introduction should clearly state the claims made, including the contributions made in the paper and important assumptions and limitations. A \answerNo{} or \answerNA{} answer to this question will not be perceived well by the reviewers. 
        \item The claims made should match theoretical and experimental results, and reflect how much the results can be expected to generalize to other settings. 
        \item It is fine to include aspirational goals as motivation as long as it is clear that these goals are not attained by the paper. 
    \end{itemize}

\item {\bf Limitations}
    \item[] Question: Does the paper discuss the limitations of the work performed by the authors?
    \item[] Answer: \answerYes{}{} 
    \item[] Justification: The paper discusses practical limitations of DyMoS, including the trade-off between motion dynamics and appearance consistency when the modulation strength is set too high.
    \item[] Guidelines:
    \begin{itemize}
        \item The answer \answerNA{} means that the paper has no limitation while the answer \answerNo{} means that the paper has limitations, but those are not discussed in the paper. 
        \item The authors are encouraged to create a separate ``Limitations'' section in their paper.
        \item The paper should point out any strong assumptions and how robust the results are to violations of these assumptions (e.g., independence assumptions, noiseless settings, model well-specification, asymptotic approximations only holding locally). The authors should reflect on how these assumptions might be violated in practice and what the implications would be.
        \item The authors should reflect on the scope of the claims made, e.g., if the approach was only tested on a few datasets or with a few runs. In general, empirical results often depend on implicit assumptions, which should be articulated.
        \item The authors should reflect on the factors that influence the performance of the approach. For example, a facial recognition algorithm may perform poorly when image resolution is low or images are taken in low lighting. Or a speech-to-text system might not be used reliably to provide closed captions for online lectures because it fails to handle technical jargon.
        \item The authors should discuss the computational efficiency of the proposed algorithms and how they scale with dataset size.
        \item If applicable, the authors should discuss possible limitations of their approach to address problems of privacy and fairness.
        \item While the authors might fear that complete honesty about limitations might be used by reviewers as grounds for rejection, a worse outcome might be that reviewers discover limitations that aren't acknowledged in the paper. The authors should use their best judgment and recognize that individual actions in favor of transparency play an important role in developing norms that preserve the integrity of the community. Reviewers will be specifically instructed to not penalize honesty concerning limitations.
    \end{itemize}

\item {\bf Theory assumptions and proofs}
    \item[] Question: For each theoretical result, does the paper provide the full set of assumptions and a complete (and correct) proof?
    \item[] Answer: \answerYes{}{} 
    \item[] Justification: Based on the observation presented in Sec.~\ref{sec:sa_comparison}, we detail our hypothesis and provide detailed validation in Sec.~\ref{sec:gamma_sweep}.
    \item[] Guidelines:
    \begin{itemize}
        \item The answer \answerNA{} means that the paper does not include theoretical results. 
        \item All the theorems, formulas, and proofs in the paper should be numbered and cross-referenced.
        \item All assumptions should be clearly stated or referenced in the statement of any theorems.
        \item The proofs can either appear in the main paper or the supplemental material, but if they appear in the supplemental material, the authors are encouraged to provide a short proof sketch to provide intuition. 
        \item Inversely, any informal proof provided in the core of the paper should be complemented by formal proofs provided in appendix or supplemental material.
        \item Theorems and Lemmas that the proof relies upon should be properly referenced. 
    \end{itemize}

    \item {\bf Experimental result reproducibility}
    \item[] Question: Does the paper fully disclose all the information needed to reproduce the main experimental results of the paper to the extent that it affects the main claims and/or conclusions of the paper (regardless of whether the code and data are provided or not)?
    \item[] Answer: \answerYes{}{} 
    \item[] Justification: We provide the complete DyMoS sampling algorithm together with detailed experimental configurations in the Appendix. We release the code to reproduce the results upon completion of the blind review period.
    \item[] Guidelines:
    \begin{itemize}
        \item The answer \answerNA{} means that the paper does not include experiments.
        \item If the paper includes experiments, a \answerNo{} answer to this question will not be perceived well by the reviewers: Making the paper reproducible is important, regardless of whether the code and data are provided or not.
        \item If the contribution is a dataset and\slash or model, the authors should describe the steps taken to make their results reproducible or verifiable. 
        \item Depending on the contribution, reproducibility can be accomplished in various ways. For example, if the contribution is a novel architecture, describing the architecture fully might suffice, or if the contribution is a specific model and empirical evaluation, it may be necessary to either make it possible for others to replicate the model with the same dataset, or provide access to the model. In general. releasing code and data is often one good way to accomplish this, but reproducibility can also be provided via detailed instructions for how to replicate the results, access to a hosted model (e.g., in the case of a large language model), releasing of a model checkpoint, or other means that are appropriate to the research performed.
        \item While NeurIPS does not require releasing code, the conference does require all submissions to provide some reasonable avenue for reproducibility, which may depend on the nature of the contribution. For example
        \begin{enumerate}
            \item If the contribution is primarily a new algorithm, the paper should make it clear how to reproduce that algorithm.
            \item If the contribution is primarily a new model architecture, the paper should describe the architecture clearly and fully.
            \item If the contribution is a new model (e.g., a large language model), then there should either be a way to access this model for reproducing the results or a way to reproduce the model (e.g., with an open-source dataset or instructions for how to construct the dataset).
            \item We recognize that reproducibility may be tricky in some cases, in which case authors are welcome to describe the particular way they provide for reproducibility. In the case of closed-source models, it may be that access to the model is limited in some way (e.g., to registered users), but it should be possible for other researchers to have some path to reproducing or verifying the results.
        \end{enumerate}
    \end{itemize}

\item {\bf Open access to data and code}
    \item[] Question: Does the paper provide open access to the data and code, with sufficient instructions to faithfully reproduce the main experimental results, as described in supplemental material?
    \item[] Answer: \answerYes{}{} 
    \item[] Justification: We provide the pseudo-code of DyMos in Appendix.~\ref{app:algorithm} for reproducibility. Additionally, we publicly release our codes and instructions as stated above.
    \item[] Guidelines:
    \begin{itemize}
        \item The answer \answerNA{} means that paper does not include experiments requiring code.
        \item Please see the NeurIPS code and data submission guidelines (\url{https://neurips.cc/public/guides/CodeSubmissionPolicy}) for more details.
        \item While we encourage the release of code and data, we understand that this might not be possible, so \answerNo{} is an acceptable answer. Papers cannot be rejected simply for not including code, unless this is central to the contribution (e.g., for a new open-source benchmark).
        \item The instructions should contain the exact command and environment needed to run to reproduce the results. See the NeurIPS code and data submission guidelines (\url{https://neurips.cc/public/guides/CodeSubmissionPolicy}) for more details.
        \item The authors should provide instructions on data access and preparation, including how to access the raw data, preprocessed data, intermediate data, and generated data, etc.
        \item The authors should provide scripts to reproduce all experimental results for the new proposed method and baselines. If only a subset of experiments are reproducible, they should state which ones are omitted from the script and why.
        \item At submission time, to preserve anonymity, the authors should release anonymized versions (if applicable).
        \item Providing as much information as possible in supplemental material (appended to the paper) is recommended, but including URLs to data and code is permitted.
    \end{itemize}

\item {\bf Experimental setting/details}
    \item[] Question: Does the paper specify all the training and test details (e.g., data splits, hyperparameters, how they were chosen, type of optimizer) necessary to understand the results?
    \item[] Answer: \answerYes{} 
    \item[] Justification: We include all evaluation methods in the main paper and Appendix. We will include them with the code release as stated above.
    \item[] Guidelines:
    \begin{itemize}
        \item The answer \answerNA{} means that the paper does not include experiments.
        \item The experimental setting should be presented in the core of the paper to a level of detail that is necessary to appreciate the results and make sense of them.
        \item The full details can be provided either with the code, in appendix, or as supplemental material.
    \end{itemize}

\item {\bf Experiment statistical significance}
    \item[] Question: Does the paper report error bars suitably and correctly defined or other appropriate information about the statistical significance of the experiments?
    \item[] Answer: \answerNo{}{} 
    \item[] Justification: The paper reports results across multiple backbones and datasets, but does not include error bars, confidence intervals, or statistical significance tests for the main quantitative results due to high computational cost on multi-seed generation.
    \item[] Guidelines:
    \begin{itemize}
        \item The answer \answerNA{} means that the paper does not include experiments.
        \item The authors should answer \answerYes{} if the results are accompanied by error bars, confidence intervals, or statistical significance tests, at least for the experiments that support the main claims of the paper.
        \item The factors of variability that the error bars are capturing should be clearly stated (for example, train/test split, initialization, random drawing of some parameter, or overall run with given experimental conditions).
        \item The method for calculating the error bars should be explained (closed form formula, call to a library function, bootstrap, etc.)
        \item The assumptions made should be given (e.g., Normally distributed errors).
        \item It should be clear whether the error bar is the standard deviation or the standard error of the mean.
        \item It is OK to report 1-sigma error bars, but one should state it. The authors should preferably report a 2-sigma error bar than state that they have a 96\% CI, if the hypothesis of Normality of errors is not verified.
        \item For asymmetric distributions, the authors should be careful not to show in tables or figures symmetric error bars that would yield results that are out of range (e.g., negative error rates).
        \item If error bars are reported in tables or plots, the authors should explain in the text how they were calculated and reference the corresponding figures or tables in the text.
    \end{itemize}

\item {\bf Experiments compute resources}
    \item[] Question: For each experiment, does the paper provide sufficient information on the computer resources (type of compute workers, memory, time of execution) needed to reproduce the experiments?
    \item[] Answer: \answerYes{} 
    \item[] Justification: Yes, please refer to Appendix.~\ref{app:impl}.
    \item[] Guidelines:
    \begin{itemize}
        \item The answer \answerNA{} means that the paper does not include experiments.
        \item The paper should indicate the type of compute workers CPU or GPU, internal cluster, or cloud provider, including relevant memory and storage.
        \item The paper should provide the amount of compute required for each of the individual experimental runs as well as estimate the total compute. 
        \item The paper should disclose whether the full research project required more compute than the experiments reported in the paper (e.g., preliminary or failed experiments that didn't make it into the paper). 
    \end{itemize}
    
\item {\bf Code of ethics}
    \item[] Question: Does the research conducted in the paper conform, in every respect, with the NeurIPS Code of Ethics \url{https://neurips.cc/public/EthicsGuidelines}?
    \item[] Answer: \answerYes{} 
    \item[] Justification: We reviewed the NeurIPS Code of Ethics.
    \item[] Guidelines:
    \begin{itemize}
        \item The answer \answerNA{} means that the authors have not reviewed the NeurIPS Code of Ethics.
        \item If the authors answer \answerNo, they should explain the special circumstances that require a deviation from the Code of Ethics.
        \item The authors should make sure to preserve anonymity (e.g., if there is a special consideration due to laws or regulations in their jurisdiction).
    \end{itemize}

\item {\bf Broader impacts}
    \item[] Question: Does the paper discuss both potential positive societal impacts and negative societal impacts of the work performed?
    \item[] Answer: \answerNo{} 
    \item[] Justification: Our work currently focuses primarily on the motion dynamics of the videos, which is not directly associated with negative societal impacts. However, if reviewers find it necessary, we will include a discussion for potential harms with generative models, particularly regarding the controllable generation of motion dynamics.
    \item[] Guidelines:
    \begin{itemize}
        \item The answer \answerNA{} means that there is no societal impact of the work performed.
        \item If the authors answer \answerNA{} or \answerNo, they should explain why their work has no societal impact or why the paper does not address societal impact.
        \item Examples of negative societal impacts include potential malicious or unintended uses (e.g., disinformation, generating fake profiles, surveillance), fairness considerations (e.g., deployment of technologies that could make decisions that unfairly impact specific groups), privacy considerations, and security considerations.
        \item The conference expects that many papers will be foundational research and not tied to particular applications, let alone deployments. However, if there is a direct path to any negative applications, the authors should point it out. For example, it is legitimate to point out that an improvement in the quality of generative models could be used to generate Deepfakes for disinformation. On the other hand, it is not needed to point out that a generic algorithm for optimizing neural networks could enable people to train models that generate Deepfakes faster.
        \item The authors should consider possible harms that could arise when the technology is being used as intended and functioning correctly, harms that could arise when the technology is being used as intended but gives incorrect results, and harms following from (intentional or unintentional) misuse of the technology.
        \item If there are negative societal impacts, the authors could also discuss possible mitigation strategies (e.g., gated release of models, providing defenses in addition to attacks, mechanisms for monitoring misuse, mechanisms to monitor how a system learns from feedback over time, improving the efficiency and accessibility of ML).
    \end{itemize}
    
\item {\bf Safeguards}
    \item[] Question: Does the paper describe safeguards that have been put in place for responsible release of data or models that have a high risk for misuse (e.g., pre-trained language models, image generators, or scraped datasets)?
    \item[] Answer: \answerNA{} 
    \item[] Justification: The paper does not release a new pretrained generative model, model checkpoint, or scraped dataset. DyMoS is an inference-time method applied to existing I2V backbones, and any code release will follow the licenses and usage restrictions of those underlying models.
    \item[] Guidelines:
    \begin{itemize}
        \item The answer \answerNA{} means that the paper poses no such risks.
        \item Released models that have a high risk for misuse or dual-use should be released with necessary safeguards to allow for controlled use of the model, for example by requiring that users adhere to usage guidelines or restrictions to access the model or implementing safety filters. 
        \item Datasets that have been scraped from the Internet could pose safety risks. The authors should describe how they avoided releasing unsafe images.
        \item We recognize that providing effective safeguards is challenging, and many papers do not require this, but we encourage authors to take this into account and make a best faith effort.
    \end{itemize}

\item {\bf Licenses for existing assets}
    \item[] Question: Are the creators or original owners of assets (e.g., code, data, models), used in the paper, properly credited and are the license and terms of use explicitly mentioned and properly respected?
    \item[] Answer: \answerYes{} 
    \item[] Justification: Yes, the paper cites the original paper that produced dataset and models.
    \item[] Guidelines:
    \begin{itemize}
        \item The answer \answerNA{} means that the paper does not use existing assets.
        \item The authors should cite the original paper that produced the code package or dataset.
        \item The authors should state which version of the asset is used and, if possible, include a URL.
        \item The name of the license (e.g., CC-BY 4.0) should be included for each asset.
        \item For scraped data from a particular source (e.g., website), the copyright and terms of service of that source should be provided.
        \item If assets are released, the license, copyright information, and terms of use in the package should be provided. For popular datasets, \url{paperswithcode.com/datasets} has curated licenses for some datasets. Their licensing guide can help determine the license of a dataset.
        \item For existing datasets that are re-packaged, both the original license and the license of the derived asset (if it has changed) should be provided.
        \item If this information is not available online, the authors are encouraged to reach out to the asset's creators.
    \end{itemize}

\item {\bf New assets}
    \item[] Question: Are new assets introduced in the paper well documented and is the documentation provided alongside the assets?
    \item[] Answer: \answerNA{} 
    \item[] Justification: The paper does not release new assets. 
    \item[] Guidelines:
    \begin{itemize}
        \item The answer \answerNA{} means that the paper does not release new assets.
        \item Researchers should communicate the details of the dataset\slash code\slash model as part of their submissions via structured templates. This includes details about training, license, limitations, etc. 
        \item The paper should discuss whether and how consent was obtained from people whose asset is used.
        \item At submission time, remember to anonymize your assets (if applicable). You can either create an anonymized URL or include an anonymized zip file.
    \end{itemize}

\item {\bf Crowdsourcing and research with human subjects}
    \item[] Question: For crowdsourcing experiments and research with human subjects, does the paper include the full text of instructions given to participants and screenshots, if applicable, as well as details about compensation (if any)? 
    \item[] Answer: \answerYes{} 
    \item[] Justification: The paper includes MTurk user study. The main paper describes the blinded and randomized comparison protocol, the evaluated criteria, and the number of image--prompt pairs and responses, while Appendix~\ref{app:user_study} provides the participant-facing questions and detailed evaluation protocol. Method names are hidden, video order is randomized, and results are reported only as aggregate preference fractions.
    \item[] Guidelines:
    \begin{itemize}
        \item The answer \answerNA{} means that the paper does not involve crowdsourcing nor research with human subjects.
        \item Including this information in the supplemental material is fine, but if the main contribution of the paper involves human subjects, then as much detail as possible should be included in the main paper. 
        \item According to the NeurIPS Code of Ethics, workers involved in data collection, curation, or other labor should be paid at least the minimum wage in the country of the data collector. 
    \end{itemize}

\item {\bf Institutional review board (IRB) approvals or equivalent for research with human subjects}
    \item[] Question: Does the paper describe potential risks incurred by study participants, whether such risks were disclosed to the subjects, and whether Institutional Review Board (IRB) approvals (or an equivalent approval/review based on the requirements of your country or institution) were obtained?
    \item[] Answer: \answerNo{} 
    \item[] Justification: The paper includes an MTurk preference study, but no IRB approval, exemption, or equivalent institutional review was obtained. The study involves low-risk video preference judgments and reports only aggregate results.
    \item[] Guidelines:
    \begin{itemize}
        \item The answer \answerNA{} means that the paper does not involve crowdsourcing nor research with human subjects.
        \item Depending on the country in which research is conducted, IRB approval (or equivalent) may be required for any human subjects research. If you obtained IRB approval, you should clearly state this in the paper. 
        \item We recognize that the procedures for this may vary significantly between institutions and locations, and we expect authors to adhere to the NeurIPS Code of Ethics and the guidelines for their institution. 
        \item For initial submissions, do not include any information that would break anonymity (if applicable), such as the institution conducting the review.
    \end{itemize}

\item {\bf Declaration of LLM usage}
    \item[] Question: Does the paper describe the usage of LLMs if it is an important, original, or non-standard component of the core methods in this research? Note that if the LLM is used only for writing, editing, or formatting purposes and does \emph{not} impact the core methodology, scientific rigor, or originality of the research, declaration is not required.
    \item[] Answer: \answerNA{} 
    \item[] Justification: The paper does not involve LLMs as part of proposed method, experiments, or core contributions. 
    \item[] Guidelines:
    \begin{itemize}
        \item The answer \answerNA{} means that the core method development in this research does not involve LLMs as any important, original, or non-standard components.
        \item Please refer to our LLM policy in the NeurIPS handbook for what should or should not be described.
    \end{itemize}

\end{enumerate}

\end{document}